\documentclass[pdflatex,sn-mathphys-num]{sn-jnl}
\usepackage{graphicx}%
\usepackage{multirow}%
\usepackage{amsmath,amssymb,amsfonts}%
\usepackage{amsthm}%
\usepackage{mathrsfs}%
\usepackage[title]{appendix}%
\usepackage{xcolor}%
\usepackage{textcomp}%
\usepackage{manyfoot}%
\usepackage{booktabs}%
\usepackage{algorithm}%
\usepackage{algorithmicx}%
\usepackage{algpseudocode}%
\usepackage{listings}%
\usepackage{pgfplots,url}
\usepgfplotslibrary{groupplots,fillbetween}
\pgfplotsset{compat=1.18}
\newcommand{\E}{\mathbb{E}}
\newcommand{\p}{\mathbb{P}}
\newtheorem{assumption}{Assumption}
\theoremstyle{thmstyleone}%
\newtheorem{theorem}{Theorem}%  meant for continuous numbers
\theoremstyle{thmstyleone}%
\newtheorem{proposition}{Proposition}% 

\theoremstyle{thmstyletwo}%
\newtheorem{example}{Example}%
\theoremstyle{thmstyletwo}%
\newtheorem{remark}{Remark}%

\theoremstyle{thmstyletwo}%

\theoremstyle{thmstylethree}%
\newtheorem{definition}{Definition}%

\theoremstyle{thmstylethree}%
\newtheorem{lemma}{Lemma}%

\begin{document}

\title[Article Title]{Entropy-Regularized Reinforcement Learning for Regime-Switching Stochastic Control: A Study on Distributional Classes}

%%=============================================================%%
%% GivenName	-> \fnm{Joergen W.}
%% Particle	-> \spfx{van der} -> surname prefix
%% FamilyName	-> \sur{Ploeg}
%% Suffix	-> \sfx{IV}
%% \author*[1,2]{\fnm{Joergen W.} \spfx{van der} \sur{Ploeg} 
%%  \sfx{IV}}\email{iauthor@gmail.com}
%%=============================================================%%
%%==================================%%
\author[1]{\fnm{Congde} \sur{Hu}}\email{hcd.math@gmail.com}

\author*[1]{\fnm{Danping} \sur{Li}}\email{dpli@fem.ecnu.edu.cn}

\author[2]{\fnm{Xun} \sur{Li}}\email{malixun@polyu.edu.hk}

\author[2]{\fnm{Yijie} \sur{Zeng}}\email{yi-jie.zeng@connect.polyu.hk}

\affil*[1]{\orgdiv{School of Statistics, KLATASDS-MOE, China Inclusive Ageing Finance Research Center}, \orgname{East China Normal University}, \orgaddress{\city{Shanghai}, \postcode{200062}, \state{Shanghai}, \country{China}}}

\affil[2]{\orgdiv{Department of Applied Mathematics}, \orgname{The Hong Kong Polytechnic University}, \orgaddress{\city{Kowloon}, \state{Hong Kong}, \country{China}}}

%% Sample for unstructured abstract %%
%%==================================%%
\abstract{We study entropy-regularized reinforcement learning for
finite-horizon stochastic control with Markov regime switching,
emphasizing the benefits of distributional structure priors.
The action domain and the action dependence of the exploratory
Hamiltonian identify explicit policy families, including Gaussian,
Gamma, Beta, Dirichlet, Weibull, Pareto, and log-normal distributions.
These analytical relations guide a structure-informed continuous-time
$q$-learning method: the policy respects the prescribed action support,
shares parameters with a normalized $q$-function, and incorporates
known distributional parameters before learning. This construction
avoids numerical action-space normalization and reduces the number
of parameters requiring estimation. In a consumption-investment
application, finiteness of the optimized HJB Hamiltonian implies
positive marginal value and strictly negative wealth curvature
for any smooth finite-valued solution, yielding independent Gamma
consumption and Gaussian investment policies. A logarithmic
terminal-utility benchmark permits independent assessment of
learning accuracy and discretization error. Across ten paired
random seeds, fixing the known consumption-shape parameter reduces
the mean number of updates required to reach a specified
policy-error threshold by 12.6\%, with comparable final accuracy.
A separate mesh study identifies first-order discretization error.
These results illustrate how distributional information can
simplify policy learning while retaining flexibility in the
unknown parameter functions.}

%%================================%%
%% Sample for structured abstract %%
%%================================%%


%%\pacs[JEL Classification]{D8, H51}

%%\pacs[MSC Classification]{35A01, 65L10, 65L12, 65L20, 65L70}

\maketitle
\section{Introduction}\label{sec1}
Reinforcement learning (RL) provides a framework for improving sequential decisions through interactions with an uncertain environment; see \citet{watkins1992q} and \citet{sutton2018reinforcement}. In continuous control problems, learning requires both selecting desirable actions and deciding how to explore alternatives. Entropy regularization incorporates this trade-off into the objective by rewarding randomized policies. Its practical importance is illustrated by soft actor-critic \citep{haarnoja2018soft}, while \citet{geist2019theory} provide a general theory of regularized Markov decision processes. The effectiveness of exploration also depends on the distribution used to represent a policy. Consumption must remain positive, allocation weights may lie on a simplex, and other controls may be bounded or unrestricted. A useful policy representation should reflect these domains and the way actions affect rewards and state dynamics.

Continuous-time RL connects policy randomization with stochastic control. \citet{wang2020reinforcement} formulate exploration through probability distributions over controls, average the drift and covariance of the state process, and derive an exploratory Hamilton--Jacobi--Bellman (HJB) equation. Their linear-quadratic setting yields Gaussian exploration, giving a mathematical explanation for a widely used policy family. \citet{tang2022exploratory} study the well-posedness of exploratory HJB equations and their convergence to classical control problems as exploration vanishes. For learning from observations, \citet{jia2022policy} develop martingale-based policy evaluation and temporal-difference methods, and \citet{jia2022gradient} establish continuous-time policy-gradient and actor-critic formulations. The instantaneous $q$-function of \citet{jia2023q} further supports policy improvement without explicitly estimating the state generator. These developments provide a foundation for combining analytical information about a control problem with data-driven estimation of its unknown quantities.

Financial applications demonstrate the scope of this approach. \citet{dai2023equilibrium} study exploratory equilibrium strategies for time-inconsistent mean-variance portfolio selection, and \citet{dai2026merton} use Gaussian policy randomization to learn Merton's strategies in an incomplete market. \citet{bo2025tracking} develop continuous-time $q$-learning for an optimal tracking portfolio problem with reflected state dynamics. Regime switching introduces a further challenge: market conditions can change abruptly, and present decisions depend on the possibility of future transitions. \citet{wu2024regime} address mean-variance portfolio selection with unobservable regimes by combining filtering and actor-critic learning. Related extensions include jump-diffusion models in \citet{gao2024reinforcement} and Markov regime-switching systems under Tsallis entropy in \citet{zhang2026continuous}. In our setting, the regime is observed and the transition generator is independent of the action. Switching couples the value functions across regimes, while the pointwise action maximization retains a tractable structure.

Within this literature, the passage from an exploratory HJB equation to an implementable policy representation deserves particular attention. Under Shannon differential entropy, the maximizing density has a Gibbs form whenever its normalizing integral is finite. This formula alone does not specify how to sample actions efficiently, evaluate their probabilities, or determine which parameters require estimation. Gaussian policies are natural for quadratic action dependence on an unrestricted domain, but other Hamiltonians generate asymmetric, bounded, or heavy-tailed distributions. For bounded controls, \citet{chou2017improving} demonstrate the relevance of Beta policies for bias and empirical learning performance. This motivates a systematic question: how can the action domain and Hamiltonian structure be translated into distributional information that directly guides continuous-time learning?

We address this question through a \emph{distributional structure prior}: knowledge of the policy family, its support, and parameter relations implied by the control problem. The prior is used in the parameterization, without introducing a Bayesian prior or changing the Shannon-entropy objective. In particular, knowing the functional dependence on an action does not require knowing all numerical coefficients of the state dynamics. A logarithmic-linear action term can identify a Gamma family even when its rate depends on an unknown value derivative; a negative quadratic term can identify a Gaussian family while leaving its mean and variance to be learned. Thus analytical structure supplies part of the policy specification, and observations determine the remaining parameter functions.

This distributional information offers three concrete benefits. First, matching the support of the density to the action domain ensures that sampled actions satisfy the corresponding pointwise constraints without an additional clipping step. Second, explicit normalized densities permit direct sampling and log-density evaluation. Defining the candidate optimal $q$-function as $\lambda$ times the policy log-density, where $\lambda>0$ is the entropy temperature, enforces its normalization identity throughout learning and links the action-dependent coefficients to the normalizing term. This avoids a separate numerical integration over the action space. Third, known shape parameters and other structural relations can be imposed exactly, reducing the number of quantities estimated from noisy observations. When these restrictions are correct, they remove redundant estimation while retaining the target policy within the chosen family. The resulting parameters also retain an interpretation in terms of location, dispersion, skewness, or tail behavior. State-process admissibility still requires the appropriate regularity and integrability conditions, beyond having a valid density.

Existing convergence analyses help place these benefits in perspective. \citet{giegrich2024convergence} establish convergence of policy-gradient methods for exploratory linear-quadratic control, and \citet{szpruch2024scheduling} analyze entropy scheduling and regret in continuous-time linear-quadratic learning. Our emphasis is on using distributional structure to simplify the learning representation across several action domains, with a fixed temperature during each training run. Implementation also requires distinguishing averaged exploratory dynamics from actions sampled on a time grid, as emphasized by \citet{jia2025erratum} and \citet{jia2026accuracy}. We therefore assess the contribution of a known distributional parameter separately from the error introduced by time discretization.

The paper makes three contributions. First, we identify explicit exploratory policy families within a common finite-horizon regime-switching framework, including Gaussian, Exponential, Gamma, Beta, Dirichlet, Weibull, Pareto, and log-normal distributions. The derivations relate each family to its action domain and Hamiltonian terms, and specify the sign and integrability restrictions needed for normalization. Second, we incorporate these relations into a structure-informed $q$-learning method. The policy and normalized $q$-function share parameters, and known distributional quantities are fixed before estimation. The learner uses observed states, regimes, actions, and rewards; the numerical drift, volatility, and regime-transition coefficients need not be supplied, although the structural action dependence is assumed known.

Third, we develop a consumption-investment application building on \citet{merton1971optimum} and \citet{sotomayor2009explicit}. Positive consumption and an unrestricted investment proportion produce independent Gamma and Gaussian policies. For any smooth finite-valued solution of the exploratory HJB equation, finiteness of its optimized Hamiltonian implies positive marginal value and strictly negative wealth curvature, establishing the signs required by both densities. A logarithmic terminal-utility benchmark gives an explicit wealth coefficient and a linear system for regime-dependent value offsets. It also identifies a consumption shape parameter determined entirely by the discount rate and entropy temperature. In ten paired numerical runs, fixing this known shape reduces the mean number of updates needed to reach a prescribed policy-error threshold by 12.6\%, with comparable final accuracy. This comparison isolates a practical benefit of the distributional prior within the same Gamma--Gaussian family. 

The remainder of the paper is organized as follows. Section~\ref{Section 2} formulates the exploratory control problem. Section~\ref{Section 3} derives the distributional classes, and Section~\ref{Section 4} develops the learning method. Section~\ref{Section 5} presents the consumption-investment application and numerical experiments. Section~\ref{sec:conclusion} concludes, and the appendices provide supporting derivations and proofs.

\textbf{Notations:} Throughout the paper, by convention, all vectors are column vectors unless otherwise specified. 
Let $\mathbb R_{> 0}$, $\mathbb R_{\ge 0}$, $\mathbb R_{< 0}$, and $\mathbb R_{\le 0}$ denote the strictly positive, non-negative, strictly negative, and non-positive reals, respectively, and $\mathbb R_K^d:=\{x\in\mathbb R^d:\lvert x\rvert\le K\}$. 
Let $\mathbb R^{k\times\ell}$ be the space of real $k\times\ell$ matrices. For a matrix $A$, $A^\top$ is its transpose, 
$\lvert A\rvert$ its Frobenius norm, and we write $A^2 := AA^\top$. Given two matrices $A$ and $B$ of the same size, 
$A\circ B=\mathrm{tr}(AB^\top)$. For a positive definite matrix $A$, we denote by $A^{1/2} = VD^{1/2}V^\top$ its principal square root matrix, where $A = VDV^\top$ is the eigendecomposition with $V$ as an orthogonal matrix and $D$ as a diagonal matrix containing the eigenvalues of $A$, and $D^{1/2}$ is the diagonal matrix whose entries are the square root of those of $D$.
For a function $f$, $f_x$ and $f_{xx}$ denote the first- and second-order partial derivatives. 
The symbol $\mathbb E_{t,x,i}[\cdot]$ denotes the conditional expectation given $X_t=x$, $\theta_t=i$. 
The symbol $\delta$ denotes the variational operator. 
Finally, $U(B)$ denotes the uniform distribution on $B$, $\mathcal N(\cdot \mid\mu,\Sigma)$ a Gaussian distribution, and $\Phi$ the cdf of $\mathcal N(0,1)$.
\section{Formulation and problem}\label{Section 2}
This section develops the mathematical framework for a stochastic optimal control problem within a regime-switching diffusion model incorporating learning. To facilitate comparison between the cases with and without learning, we adopt as a benchmark the classical optimal control problem, outlined in Subsection \ref{sub21}. Finally, Subsection \ref{sub23} provides the integrated model of the optimal control problem with ERRL.

\subsection{Classical optimal control within regime-switching diffusion model}\label{sub21}
For a fixed positive constant $T$ and $t\in[0,T]$, let $\left(\Omega_t, \mathcal{F}_t,\{\mathcal{F}_{t,s}\}_{t\leq s \leq T}, \p_t\right)$
be a filtered probability space satisfying the usual conditions. Let $\theta = \{\theta_s, t \leq s \leq T\}$ be a continuous-time Markov chain on this probability space, taking values in a finite state space $\mathcal{S} = \{1, 2, \ldots, m\}$, which represents the regime states. The generator matrix of this Markov chain is defined as $Q = (q_{ij})_{m \times m}$, where $q_{ij} \geq 0$ ($i \neq j$) represents the transition rate from regime $i$ to regime $j$, satisfying $q_{ii} = -\sum_{j=1, j \neq i}^{m} q_{ij}$. The optimal control problem is defined on $\left(\Omega_t, \mathcal{F}_t,\{\mathcal{F}_{t,s}\}_{t\leq s \leq T}, \p_t\right)$, and consists of the following controlled SDE
\begin{equation}\label{CLSDE}
\left\{\begin{array}{ll}
d X_s=b(s, X_{s}, \theta_{s} ; a_s) d s+\sigma(s, X_{s}, \theta_{s} ; a_s) d W_s, \quad s \in(t, T],\\
X_t=x, \quad \theta_t=i,
\end{array} \right.
\end{equation}
as well as the payoff function
\begin{equation}\label{CLpayoff function}
J^{cl}(t, x, i ; a)
=\E_{t,x,i}\left[\int_t^T r(s, X_s, \theta_s ; a_s) d s+g(X_T, \theta_T)\right] ,
\end{equation}
where
\begin{equation*}
\begin{aligned}
b:&[0,T]\times\mathbb{R}^d\times\mathcal{S}\times\mathcal{A}\rightarrow\mathbb{R}^d,\\
\sigma:&[0,T]\times\mathbb{R}^d\times\mathcal{S}\times\mathcal{A}\rightarrow\mathbb{R}^{d \times k},\\
r:&[0,T]\times\mathbb{R}^d\times\mathcal{S}\times\mathcal{A}\rightarrow\mathbb{R},\quad g:\mathbb{R}^d\times\mathcal{S}\rightarrow\mathbb{R},
\end{aligned}
\end{equation*}
$a$ is a predictable control process with values in a metric space $\mathcal{A}\subseteq \mathbb{R}^n$. $W_s$ is a $k$-dimensional Brownian motion on $\left(\Omega_t, \mathcal{F}_t,\{\mathcal{F}_{t,s}\}_{t\leq s \leq T}, \p_t\right)$.

For a function $h(x)$ on $\mathbb{R}^d$ that satisfies local Lipschitz continuity and linear growth, its Lipschitz norm $||h||_K$ is defined as
\begin{equation}
||h||_K := \sup_{x \in \mathbb{R}^d} \frac{|h(x)|}{1+|x|} + \sup_{\substack{x, y \in \mathbb{R}_K^d; x \neq y}} \frac{|h(x) - h(y)|}{|x - y|}.
\end{equation}

\begin{assumption}\label{Assumption 1} The following conditions hold.
\begin{itemize}
\item[$(\romannumeral 1)$] For each $i\in\mathcal S$, the functions
$b$, $\sigma$, and $r$ are continuous in $(t,x,a)$, and $g(\cdot,i)$
is continuous.
\item[$(\romannumeral 2)$] For every $K>0$ and every compact set
$B\subseteq\mathcal A$, there exists $C_{K,B}>0$ such that
$
(\|b\|_K+\|\sigma\|_K)(t,\cdot,i;a)
\leq C_{K,B}, t\in[0,T],\ i\in\mathcal S,\ a\in B.
$
\item[$(\romannumeral 3)$] There exist $C_g>0$ and $p_g\geq1$ such that
$|g(x,i)|\leq C_g(1+|x|^{p_g})$. For every compact
$B\subseteq\mathcal A$, there exist $C_B>0$ and $p_B\geq1$ such that
$
|r(t,x,i;a)|\leq C_B(1+|x|^{p_B}),
 t\in[0,T],\ x\in\mathbb R^d,\ i\in\mathcal S,\ a\in B.
$
\item[$(\romannumeral 4)$] The Markov chain $\theta$ is independent of
the Brownian motion $W$.
\end{itemize}
\end{assumption}

The bounds in Assumption~\ref{Assumption 1} are uniform only on compact
subsets of the action space. In the classical problem, an admissible control
is a predictable action process for which \eqref{CLSDE} has a unique
nonexplosive strong solution and
\[
\E_{t,x,i}\left[\int_t^T |r(s,X_s,\theta_s;a_s)|\,ds
                  +|g(X_T,\theta_T)|\right]<\infty.
\]
In particular, these properties hold for controls taking values in a
fixed compact subset of $\mathcal A$. 

In this optimal control framework, the controller aims to maximize the payoff function $J^{cl}$ by choosing an admissible control process $a$. At any time $s \in (t, T)$, the controller makes decisions based on the available information. By establishing the Dynamic Programming Principle (DPP) for this problem, one can derive the corresponding Hamilton-Jacobi-Bellman (HJB) equation.
\subsection{Entropy regularized exploratory optimal control framework}\label{sub23}
A key concept in RL is exploring the unknown environment through randomized actions. Randomization employs probability distributions (measures) to substitute deterministic actions for the purpose of exploration. Entropy is used to quantify the dispersion of these probability distributions, where low entropy indicates concentration and high entropy signifies greater randomness. This paper focuses  on feedback distributional controls, i.e., the current policy depends only on the current state $t, x$ and the regime state $i$.

Let $\boldsymbol{\pi}:[0,T]\times\mathbb{R}^d\times\mathcal{S}
\to\mathcal{P}(\mathcal{A})$ be a stochastic feedback policy, where
$\mathcal{P}(\mathcal{A})$ denotes the set of probability densities
on $\mathcal{A}$. At state $(t,x,i)$, the controller samples an action
from $\boldsymbol{\pi}(\cdot\mid t,x,i)$. Following the
generator-averaging approach of \citet{gao2024reinforcement}, we
formulate the exploratory dynamics by averaging the infinitesimal
generator of the controlled regime-switching process with respect
to this policy. The derivation is provided in
Appendix~\ref{Appendix  exploratory SDE}. The resulting regime-switching exploratory
SDE is
\begin{equation}\label{SDE}
\left\{
\begin{aligned}
dX_s
&=\tilde{b}(s,X_s,\theta_s;\boldsymbol{\pi}_s)\,ds
+\tilde{\sigma}(s,X_s,\theta_s;\boldsymbol{\pi}_s)
\,d\mathcal{W}_s,\quad s\in(t,T],\\
X_t&=x,\quad \theta_t=i,
\end{aligned}
\right.
\end{equation}
where $\mathcal{W}$ is a $d$-dimensional Brownian motion independent
of $\theta$, and
\begin{equation*}
\begin{aligned}
\tilde{b}(t,x,i;\boldsymbol{\pi})
&:=\int_{\mathcal{A}} b(t,x,i;a)
\boldsymbol{\pi}(a\mid t,x,i)\,da,\\
\tilde{\sigma}(t,x,i;\boldsymbol{\pi})
&:=\left\{
\int_{\mathcal{A}}\sigma^2(t,x,i;a)
\boldsymbol{\pi}(a\mid t,x,i)\,da
\right\}^{1/2}.
\end{aligned}
\end{equation*}

To encourage exploration, we add an entropy regularizer to the running reward of the
original control problem. The exploratory payoff function is
\begin{equation}\label{payoff function}
\begin{array}{ll}
J(t, x, i; {\boldsymbol{\pi}})
=&\E_{t,x,i}\bigg[{\displaystyle\int}_t^T \Big(\tilde{r}(s, X_{s}, \theta_s; \boldsymbol{\pi}_s)-\lambda{\displaystyle\int}_{\mathcal{A}} \boldsymbol{\pi}(a | s, X_s, \theta_s)\ln \boldsymbol{\pi}(a | s, X_s, \theta_s) \,da\Big)ds\\
&+g(X_{T}, \theta_T)\bigg],
\end{array}
\end{equation}
where
\[
\tilde{r}(t, x, i; \boldsymbol{\pi}):=\int_{\mathcal{A}} r(t, x, i; a) \boldsymbol{\pi}(a \mid t, x, i)\,da,
\]
and $\lambda>0$ is called the \textit{temperature parameter}.
\begin{definition}\label{def:admissible_policy}
(Admissible control).
A feedback density $\boldsymbol\pi$ is admissible, written
$\boldsymbol\pi\in\boldsymbol\Pi$, if the following conditions hold.
\begin{itemize}
\item[$(\romannumeral 1)$] The map $(t,x,a)\mapsto
\boldsymbol\pi(a\mid t,x,i)$ is Borel measurable for each $i$, and
$\boldsymbol\pi(\cdot\mid t,x,i)\in\mathcal P(\mathcal A)$ for every
$(t,x,i)$. Moreover,
\[
\int_{\mathcal A}\bigl(|b(t,x,i;a)|+|\sigma(t,x,i;a)|^2\bigr)
\boldsymbol\pi(a\mid t,x,i)\,da<\infty.
\]
\item[$(\romannumeral 2)$] For every $K>0$, there exists
$L_{K,\boldsymbol\pi}>0$ such that, uniformly in $t$ and $i$,
\[
\int_{\mathcal A}|\boldsymbol\pi(a\mid t,x,i)
-\boldsymbol\pi(a\mid t,y,i)|\,da
\leq L_{K,\boldsymbol\pi}|x-y|,
\quad x,y\in\mathbb R_K^d.
\]
\item[$(\romannumeral 3)$] There exist $C_{\boldsymbol\pi}>0$ and
$p_{\boldsymbol\pi}\geq2$ such that
\[
\int_{\mathcal A}\bigl(|r(t,x,i;a)|
       +|\ln\boldsymbol\pi(a\mid t,x,i)|\bigr)
\boldsymbol\pi(a\mid t,x,i)\,da
\leq C_{\boldsymbol\pi}(1+|x|^{p_{\boldsymbol\pi}}).
\]
Here $0|\ln0|$ is interpreted as zero.
\item[$(\romannumeral 4)$] For some $p\geq1$ and $C>0$,
\[
\int_{\mathcal A}|a|^p\boldsymbol\pi(a\mid t,x,i)\,da
\leq C(1+|x|^p),\quad (t,x,i)\in[0,T]\times\mathbb R^d\times\mathcal S.
\]
\item[$(\romannumeral 5)$]
For every $K>0$, there exists $C_K>0$ such that
\begin{equation}\label{eq:averaged_local_lipschitz}
|\tilde b(t,x,i;\boldsymbol\pi)
  -\tilde b(t,y,i;\boldsymbol\pi)|
 +|\tilde\sigma(t,x,i;\boldsymbol\pi)
  -\tilde\sigma(t,y,i;\boldsymbol\pi)|
\leq C_K|x-y|,
\quad x,y\in\mathbb R_K^d.
\end{equation}
There also exists $C>0$ such that
\begin{equation}\label{eq:averaged_linear_growth}
|\tilde b(t,x,i;\boldsymbol\pi)|
+|\tilde\sigma(t,x,i;\boldsymbol\pi)|
\leq C(1+|x|),
\quad x\in\mathbb R^d.
\end{equation}
\end{itemize}
\end{definition}

\begin{proposition}\label{Proposition 1}
Under Assumption \ref{Assumption 1}, for any admissible control $\boldsymbol{\pi} = \boldsymbol{\pi}(\cdot \mid t, x, i)\in\boldsymbol{\Pi}$, the exploratory SDE \eqref{SDE} has a unique strong solution.
\end{proposition}

\begin{proposition}\label{Proposition 2}
Under Assumption \ref{Assumption 1}, for any admissible control $\boldsymbol{\pi} \in \boldsymbol{\Pi}$, let $X_s^{\boldsymbol{\pi}}$ be the unique strong solution to the regime-switching exploratory SDE \eqref{SDE} with initial condition $X_t^{\boldsymbol{\pi}} = x$ and $\theta_t = i$. Then, for any $p \geq 2$, there exists a positive constant $C_{p,\boldsymbol{\pi}} > 0$, independent of $t$, $x$, and $i$, such that
\begin{equation}\label{moment_estimate}
\E_{t,x,i} \left[ \sup_{t \leq s \leq T} |X_s^{\boldsymbol{\pi}}|^p \right] \leq C_{p,\boldsymbol{\pi}}(1+|x|^p).
\end{equation}
\end{proposition}
The proofs of Propositions~\ref{Proposition 1} and~\ref{Proposition 2}
can be found in Appendices~\ref{Appendix B} and~\ref{Appendix C},
respectively.

\begin{definition}
(Value function).
The exploratory value function of the optimal control problem \eqref{SDE}-\eqref{payoff function}, with initial data $(t, x, i)$, is defined as
\begin{equation}\label{value function}
V(t, x, i):=\sup _{\boldsymbol{\pi} \in \boldsymbol{\Pi}} J(t, x, i ; \boldsymbol{\pi}).
\end{equation}
\end{definition}
Conditions~$(\romannumeral 3)$ and~$(\romannumeral 5)$ of
Definition~\ref{def:admissible_policy}, together with
Proposition~\ref{Proposition 2}, ensure that $J(t,x,i;\boldsymbol\pi)$
is finite for each admissible policy. We assume $\boldsymbol\Pi$ is
nonempty and restrict the subsequent HJB analysis to problems with finite
$V$. Whenever a Gibbs formula is used below, its normalizing integral
is assumed to be finite and strictly positive, and the resulting feedback
policy is required to belong to $\boldsymbol\Pi$. The integrability
conditions in Section~\ref{Section 3} concern this normalization;
admissibility also requires the state regularity and growth conditions
above.

The exploratory Hamilton-Jacobi-Bellman (HJB) equation associated with the value function \eqref{value function} is
\begin{equation}\label{HJB equation}
\begin{aligned}
&0=
\sup_{\boldsymbol\pi\in\boldsymbol\Pi}
\left\{\int_{\mathcal A}
\left[H(t,x,i,V_x,V_{xx};a)
-\lambda\ln\boldsymbol\pi(a\mid t,x,i)\right]
\boldsymbol\pi(a\mid t,x,i)\,da\right\}\\
&\qquad+V_t(t,x,i)+\sum_{j=1}^m q_{ij}V(t,x,j),\quad (t,x,i)\in[0,T)\times\mathbb R^d\times\mathcal S,\\
&V(T,x,i)=g(x,i),\qquad (x,i)\in\mathbb R^d\times\mathcal S.
\end{aligned}
\end{equation}
where
\[
H(t, x, i, V_x, V_{xx};a)
:=r(t, x, i;a)+b(t, x, i; a) \circ V_x + \frac{1}{2}\sigma^2(t, x, i; a) \circ V_{xx}.
\]

The goal is to establish the optimal feedback strategy $\boldsymbol{\pi}^*(a \mid t, x, i)$ that maximizes the payoff. Once $\boldsymbol{\pi}^*(a \mid t, x, i)$ is obtained, the value function can be determined by the associated PDE. Theorem \ref{Theorem Optimal strategy} below provides the optimal strategy.
\begin{theorem}\label{Theorem Optimal strategy}
(Optimal strategy).
If the HJB equation admits a solution $V(\cdot,\cdot,i) \in C^{1,2}([0, T] \times \mathbb{R}^d) \cap C([0, T] \times \mathbb{R}^d)$ for each $i \in \mathcal{S}$, then the optimal strategy $\boldsymbol{\pi}^*(a \mid t, x, i)$ is given by
\begin{equation}
\begin{aligned}
\boldsymbol{\pi}^*(a\mid t,x,i)&=\frac{\displaystyle\exp \left(\frac{1}{\lambda}H(t,x,i,V_x,V_{xx};a)\right)}{\displaystyle\int_\mathcal{A} \exp \left(\frac{1}{\lambda}H(t,x,i,V_x,V_{xx};a)\right) da}.
\end{aligned}
\end{equation}
\end{theorem}

\begin{proof}
Starting from the HJB equation, the maximizer $\boldsymbol{\pi}^*(a \mid t, x, i)$ is given by
\begin{equation*}
\arg\sup_{\boldsymbol{\pi} \in \boldsymbol{\Pi}}\left\{ \int_{\mathcal{A}} H(t, x, i, V_x, V_{xx}; a) \boldsymbol{\pi}(a \mid t, x, i)\,da - \lambda \int_{\mathcal{A}} \boldsymbol{\pi}(a \mid t, x, i) \ln \boldsymbol{\pi}(a \mid t, x, i)\,da \right\}
\end{equation*}
subject to the constraint that $\boldsymbol{\pi}$ is a probability distribution, namely
$\int_{\mathcal{A}} \boldsymbol{\pi}(a \mid t, x, i)\,da = 1.$
Introducing the Lagrange multiplier $\mu$ to this problem immediately yields
\begin{equation*}
\begin{aligned}
\mathcal{L}(\boldsymbol{\pi},\mu)=& \int_{\mathcal{A}} \Big\{ H(t, x, i, V_x, V_{xx}; a) \boldsymbol{\pi}(a \mid t, x, i)
- \lambda \boldsymbol{\pi}(a \mid t, x, i)\ln \boldsymbol{\pi}(a \mid t, x, i)\Big\} da\\
&
-\mu \left( \int_{\mathcal{A}} \boldsymbol{\pi}(a \mid t, x, i)\,da - 1 \right).
\end{aligned}
\end{equation*}
The stationary condition is
\begin{equation*}
\left\{\begin{array}{ll}
\delta \mathcal{L}&={\displaystyle\int}_{\mathcal{A}} \Big\{ H(t, x, i, V_x, V_{xx}; a) - \lambda (1 + \ln \boldsymbol{\pi}(a \mid t, x, i)) - \mu \Big\}\delta\boldsymbol{\pi} \, da=0,\\
0&={\displaystyle\int}_{\mathcal{A}} \boldsymbol{\pi}(a \mid t, x, i)\,da - 1.
\end{array} \right.
\end{equation*}
Solving the above equation for $\boldsymbol{\pi}(a \mid t, x, i)$ yields the explicit form of the maximizer $\boldsymbol{\pi}^*(a \mid t, x, i)$.
\end{proof}

\section{Several special cases}\label{Section 3}
In this section, we discuss specific forms of the optimal stochastic policy $\boldsymbol{\pi}^*(a \mid t, x, i)$ under different structural assumptions on the control space $\mathcal{A}$ and the Hamiltonian $H$. Rather than restricting the exploration noise to a standard additive Gaussian formulation, our continuous-time entropy-regularized framework reveals that the optimal exploratory distribution naturally molds itself to the underlying characteristics of the control problem. By varying the domain constraints and the algebraic structure of the Hamiltonian, we  derive a rich family of explicit exploratory distributions, including the \textbf{Gaussian, Exponential, Gamma, Beta, Dirichlet, Weibull, Pareto, and Log-Normal} distributions.

The emergence of these diverse distributions under a unified theoretical framework carries profound theoretical and practical implications for continuous-time reinforcement learning:
\begin{itemize}
    \item \textbf{Intrinsic adaptation to action space geometry:} By incorporating the topological boundaries of $\mathcal{A}$ (e.g., the non-negative orthant, bounded intervals, or the probability simplex) into the normalizability conditions of the density function, the resulting policies respect physical and operational constraints without requiring ad-hoc boundary projections or action clipping.
    \item \textbf{Structure-preserving exploration:} The mathematical form of the Hamiltonian (e.g., linear, quadratic, logarithmic, or power-law cost structures) directly dictates the shape and tail behavior of the optimal distribution. This demonstrates that optimal exploration is not arbitrary; it is  governed by the physical or economic nature of the system's intrinsic objective function.
    \item \textbf{Connection with classical control:} Whenever a modal action exists, it maximizes the Hamiltonian and therefore has the same functional form as the corresponding classical feedback rule. The mean need not have this property.
    \item \textbf{Guidance for practical algorithm design:} Deriving these explicit, closed-form probability density functions provides a  theoretical foundation for designing customized parameterized actor networks (e.g., in Soft Actor-Critic algorithms \citet{haarnoja2018soft}). Practitioners can select the specific distributional family that matches their system's structure, ensuring stable, mathematically well-posed, and domain-aware exploration in complex continuous action spaces.
\end{itemize}

For each distribution, we derive its mean, variance or covariance, and mode.
By the strict monotonicity of the exponential function,
\begin{equation}\label{eq:mode_hamiltonian}
\begin{aligned}
\mathcal{M}(t,x,i)
&:=\operatorname*{arg\,max}_{a\in\mathcal{A}}
\boldsymbol{\pi}^*(a\mid t,x,i)\\
&=\operatorname*{arg\,max}_{a\in\mathcal{A}}
H(t,x,i,V_x,V_{xx};a).
\end{aligned}
\end{equation}
Thus, the joint mode has the same functional form as the corresponding
classical optimal strategy. For open action spaces, modal locations on
excluded boundaries are understood in a limiting sense.

\subsection{Gaussian distribution}
Suppose the control space is the Euclidean space, i.e., $\mathcal{A} = \mathbb{R}^n$, and the Hamiltonian $H$ is quadratic with respect to the control variable $a$. To ensure that the resulting optimal policy forms a well-defined probability distribution (i.e., its density integrates to a finite value over the domain $\mathbb{R}^n$), we assume that the quadratic term is negative definite. Specifically, the Hamiltonian can be expressed in the following general quadratic form
\begin{equation}
\begin{aligned}
H(t,x,i,V_x,V_{xx};a)
=&r(t, x, i;a)+b(t, x, i; a) \circ V_x + \frac{1}{2}\sigma^2(t, x, i; a) \circ V_{xx}\\
=&-\frac{1}{2} a^\top R(t,x,i,V_x,V_{xx})a + L(t,x,i,V_x,V_{xx})^\top a \\
&+ C(t,x,i,V_x,V_{xx}),
\end{aligned}
\end{equation}
where $R(t,x,i,V_x,V_{xx}) \in \mathbb{R}^{n \times n}$ is a symmetric positive definite matrix-valued function, $L(t,x,i,V_x,V_{xx}) \in \mathbb{R}^n$ is a vector-valued function, and $C(t,x,i,V_x,V_{xx}) \in \mathbb{R}$ is a scalar function independent of $a$. By our preceding assumption, the matrix $R$ is positive definite, ensuring that the quadratic term $- \frac{1}{2} a^\top R a$ dominates and goes to $-\infty$ as $|a| \to \infty$. For notational simplicity, we drop the arguments and denote these terms simply as $R$, $L$, and $C$.

According to Theorem \ref{Theorem Optimal strategy}, the optimal strategy is
\begin{equation}
\boldsymbol{\pi}^*(a\mid t,x,i)
=\frac{\displaystyle\exp \left( \frac{1}{\lambda} \left( -\frac{1}{2} a^\top R a + L^\top a + C \right) \right)}{\displaystyle\int_{\mathbb{R}^n} \exp \left( \frac{1}{\lambda} \left( -\frac{1}{2} a^\top R a + L^\top a + C \right) \right) da}.
\end{equation}
The scalar constant $\exp(C/\lambda)$ appears in both the numerator and the denominator and thus cancels out. By completing the square for the terms involving $a$ in the exponent, we obtain
\begin{equation}
-\frac{1}{2} a^\top R a + L^\top a = -\frac{1}{2} \left( a - R^{-1}L \right)^\top R \left( a - R^{-1}L \right) + \frac{1}{2} L^\top R^{-1} L.
\end{equation}
Substituting this back into the policy expression, the term $\exp\left(\frac{1}{2\lambda} L^\top R^{-1} L\right)$ also cancels out as a constant. The optimal strategy then simplifies to the standard probability density function of a multivariate Gaussian distribution
\begin{equation}
\boldsymbol{\pi}^*(a\mid t,x,i) = \frac{\displaystyle\exp \left( -\frac{1}{2} \left( a - R^{-1}L \right)^\top \left( \frac{ R}{\lambda} \right) \left( a - R^{-1}L \right) \right)}{\displaystyle\int_{\mathbb{R}^n} \exp \left( -\frac{1}{2} \left( a - R^{-1}L \right)^\top \left( \frac{ R}{\lambda} \right) \left( a - R^{-1}L \right) \right) da}.
\end{equation}
Therefore, under the optimal exploratory policy, the control vector $a$ follows a multivariate Gaussian distribution
\begin{equation}
\boldsymbol{\pi}^*(\cdot \mid t,x,i) = \mathcal{N}\left(\cdot \mid R^{-1}L,\, \lambda R^{-1} \right).
\end{equation}

The mean, covariance matrix, and unique joint mode are given by
\begin{equation}
\begin{aligned}
\mathbb{E}^{\boldsymbol{\pi}^*}[a] &= R(t,x,i,V_x,V_{xx})^{-1} L(t,x,i,V_x,V_{xx}),\\
\text{Cov}^{\boldsymbol{\pi}^*}[a] &= \lambda R(t,x,i,V_x,V_{xx})^{-1},\\
a^{\mathrm{mode}} &= R(t,x,i,V_x,V_{xx})^{-1} L(t,x,i,V_x,V_{xx}).
\end{aligned}
\end{equation}

\begin{remark}
This explicit formulation connects entropy-regularized reinforcement learning with classical Linear-Quadratic (LQ) optimal control. The joint mode $a^{\mathrm{mode}}=R^{-1}L$ solves $-Ra+L=0$ and has the same functional form as the optimal deterministic feedback in standard LQ problems. For the Gaussian distribution, the mean coincides with the mode, so it shares this form as well. This correspondence does not identify the exploratory and classical value functions. With $R$ and $L$ fixed, the temperature parameter $\lambda$ scales the covariance matrix, and the policy converges to a Dirac measure at $R^{-1}L$ as $\lambda\to0^+$. A larger $\lambda$ increases the dispersion around the modal action.
\end{remark}

\subsection{Exponential distribution}\label{subsection Exponential distribution}
Suppose the control space is the non-negative orthant, i.e., $\mathcal{A}=(\mathbb{R}_{\ge 0})^n \subseteq \mathbb{R}^n$, and the Hamiltonian $H$ is linear with respect to the control variable $a$. To ensure that the resulting optimal policy forms a well-defined probability distribution (i.e., its density integrates to a finite value over the domain $\mathcal{A}$), we assume that the coefficients of this linear part are negative. Specifically, the Hamiltonian can be expressed as
\begin{equation}
\begin{aligned}
H(t,x,i,V_x,V_{xx};a)
=&r(t, x, i;a)+b(t, x, i; a) \circ V_x + \frac{1}{2}\sigma^2(t, x, i; a) \circ V_{xx}\\
=&L(t,x,i,V_x,V_{xx})^\top a+C(t,x,i,V_x,V_{xx}),
\end{aligned}
\end{equation}
where $C(t,x,i,V_x,V_{xx})\in\mathbb{R}$ is a scalar function independent of $a$. $L(t,x,i,V_x,V_{xx}) = (L_1, \dots, L_n)^\top \in (\mathbb{R}_{< 0})^n$ is a vector-valued function independent of $a$. By our preceding assumption, $L_k < 0$ for all $k=1,\dots,n$. 

According to Theorem \ref{Theorem Optimal strategy}, the explicit form of the optimal strategy is
\begin{equation}
\boldsymbol{\pi}^*(a\mid t,x,i)
=\frac{\displaystyle\exp \left(\frac{1}{\lambda} L(t,x,i,V_x,V_{xx})^\top a+\frac{1}{\lambda}C(t,x,i,V_x,V_{xx})\right)}{\displaystyle\int_{(\mathbb{R}_{\ge 0})^n} \exp \left(\frac{1}{\lambda} L(t,x,i,V_x,V_{xx})^\top a+\frac{1}{\lambda}C(t,x,i,V_x,V_{xx})\right) da}.
\end{equation}
Since $L_k < 0$ for all $k$, the multiple integral in the denominator factors into the product of $n$ convergent one-dimensional integrals. Consequently, the optimal policy factors into a product of independent one-dimensional exponential distributions
\begin{equation}
\boldsymbol{\pi}^*(a\mid t,x,i) = \prod_{k=1}^n \left( -\frac{L_k}{\lambda} \right) \exp\left( \frac{L_k}{\lambda} a_k \right).
\end{equation}

Therefore, under the optimal exploratory policy, each control component $a_k$ is independent and follows an exponential distribution with the rate parameter $-L_k/\lambda$,
\begin{equation}
a_k \sim \mathrm{Exp}\left(-\frac{L_k(t,x,i,V_x,V_{xx})}{\lambda}\right), \quad k = 1, \dots, n.
\end{equation}
The mean, variance, and mode of each component are given by
\begin{equation}
\begin{aligned}
\mathbb{E}^{\boldsymbol{\pi}^*}[a_k]
&=-\frac{\lambda}{L_k(t,x,i,V_x,V_{xx})},\\
\text{Var}^{\boldsymbol{\pi}^*}[a_k]
&=\frac{\lambda^2}{L_k(t,x,i,V_x,V_{xx})^2},\\
a_k^{\mathrm{mode}}&=0,\qquad k=1,\dots,n.
\end{aligned}
\end{equation}
The density is strictly decreasing in each component, so its unique joint
mode is $a^{\mathrm{mode}}=\mathbf{0}$.

\begin{remark}
Since $L_k<0$, the modal action $a_k^{\mathrm{mode}}=0$ maximizes the linear Hamiltonian over the non-negative half-line for every $\lambda>0$. It therefore has the same form as the classical deterministic boundary strategy. The mean $-\lambda/L_k$ is strictly positive and describes the average exploratory action. With $L_k$ fixed, both the mean and variance tend to zero as $\lambda\to0^+$, while a larger $\lambda$ spreads exploration further from the modal boundary action.
\end{remark}

\subsection{Gamma distribution}
Suppose the control space is the positive orthant, i.e., $\mathcal{A}=(\mathbb{R}_{> 0})^n \subseteq \mathbb{R}^n$, and the Hamiltonian $H$ is additively separable with respect to the control components, containing both logarithmic and linear terms. This structure frequently appears in economic and financial applications involving logarithmic utility functions. Specifically, the Hamiltonian can be expressed in vector form as
\begin{equation}
\begin{aligned}
H(t,x,i,V_x,V_{xx};a)
=&r(t, x, i;a)+b(t, x, i; a) \circ V_x + \frac{1}{2}\sigma^2(t, x, i; a) \circ V_{xx}\\
=&K(t,x,i,V_x,V_{xx})^\top \ln(a) + L(t,x,i,V_x,V_{xx})^\top a \\
&+ C(t,x,i,V_x,V_{xx}),
\end{aligned}
\end{equation}
where $K(t,x,i,V_x,V_{xx}) \in (-\lambda, \infty)^n$ is a vector-valued function whose components $K_k > -\lambda$ ensure the integrability of the probability density near the origin (as $a_k \to 0^+$), $L(t,x,i,V_x,V_{xx}) \in (\mathbb{R}_{< 0})^n$ is a vector-valued function whose negative components $L_k < 0$ guarantee the exponential tail decay of the distribution (as $a_k \to \infty$), $C(t,x,i,V_x,V_{xx}) \in \mathbb{R}$ is a scalar function, and $\ln(a) := (\ln a_1, \dots, \ln a_n)^\top$ denotes the element-wise natural logarithm of the control vector. Together, these constraints on the value spaces of $K$ and $L$ ensure that the resulting optimal policy forms a well-defined probability distribution over $(\mathbb{R}_{> 0})^n$. For notational simplicity, we drop the arguments and denote these vectors simply as $K$ and $L$.

According to Theorem \ref{Theorem Optimal strategy}, the optimal strategy is
\begin{equation}
\boldsymbol{\pi}^*(a\mid t,x,i)=\frac{\displaystyle\exp \left( \frac{1}{\lambda} K^\top \ln(a) + \frac{1}{\lambda} L^\top a+\frac{1}{\lambda}C(t,x,i,V_x,V_{xx}) \right)}{\displaystyle\int_{(\mathbb{R}_{> 0})^n} \exp \left( \frac{1}{\lambda} K^\top \ln(a) + \frac{1}{\lambda} L^\top a+\frac{1}{\lambda}C(t,x,i,V_x,V_{xx}) \right) da}.
\end{equation}
Note that the constant term $\exp(C/\lambda)$ cancels out. By expanding the dot products into sums and applying the algebraic identity $\exp(a \ln x) = x^a$, the expression factors into the product of $n$ independent one-dimensional functions
\begin{equation}
\boldsymbol{\pi}^*(a\mid t,x,i) \propto \prod_{k=1}^n a_k^{\frac{K_k}{\lambda}} \exp\left( \frac{L_k}{\lambda} a_k \right).
\end{equation}
By matching this functional form with the probability density function of a standard Gamma distribution $f(x; \alpha, \beta) \propto x^{\alpha-1} e^{-\beta x}$, we can deduce that under the optimal exploratory policy, each control component $a_k$ is independent and follows a Gamma distribution
\begin{equation}
a_k \sim \mathrm{Gamma}\left(\alpha_k, \beta_k\right), \quad k = 1, \dots, n,
\end{equation}
where the shape parameter $\alpha_k$ and the rate parameter $\beta_k$ are given by
\begin{equation}
\alpha_k = \frac{K_k(t,x,i,V_x,V_{xx})}{\lambda} + 1, \quad \beta_k = -\frac{L_k(t,x,i,V_x,V_{xx})}{\lambda}.
\end{equation}

The expected optimal control action and the variance for each component under the exploratory policy are given by
\begin{equation}
\begin{aligned}
\mathbb{E}^{\boldsymbol{\pi}^*}[a_k] &= \frac{\alpha_k}{\beta_k} = -\frac{K_k(t,x,i,V_x,V_{xx}) + \lambda}{L_k(t,x,i,V_x,V_{xx})},\\
\text{Var}^{\boldsymbol{\pi}^*}[a_k] &= \frac{\alpha_k}{\beta_k^2} = \frac{\lambda \big( K_k(t,x,i,V_x,V_{xx}) + \lambda \big)}{L_k(t,x,i,V_x,V_{xx})^2}.
\end{aligned}
\end{equation}

If $K_k>0$ (equivalently, $\alpha_k>1$), differentiating the logarithm
of the marginal density yields
\begin{equation}
\frac{d}{da_k}\ln\pi_k^*(a_k)
=\frac{\alpha_k-1}{a_k}-\beta_k
=\frac{1}{\lambda}\left(\frac{K_k}{a_k}+L_k\right).
\end{equation}
Hence the unique interior mode is
\begin{equation}
a_k^{\mathrm{mode}}=\frac{\alpha_k-1}{\beta_k}
=-\frac{K_k}{L_k},\qquad K_k>0.
\end{equation}
If $-\lambda<K_k\leq0$, the density is strictly decreasing and has a
limiting boundary peak at $a_k=0$. 

\begin{remark}
For $K_k>0$, the modal action $-K_k/L_k$ solves the Hamiltonian first-order condition $K_k/a_k+L_k=0$ and has the same functional form as the classical deterministic feedback. The mean satisfies $\mathbb{E}^{\boldsymbol{\pi}^*}[a_k]=a_k^{\mathrm{mode}}-\lambda/L_k>a_k^{\mathrm{mode}}$, so the mode, rather than the mean, gives this correspondence at a positive temperature. The coefficients still depend on the exploratory value function. With $K_k>0$ and $L_k<0$ fixed, the variance vanishes and the policy concentrates at its mode as $\lambda\to0^+$. The shape parameter $\alpha_k=K_k/\lambda+1$ shows how the temperature changes the asymmetry of exploration.
\end{remark}
\begin{remark}
The Gamma distribution framework serves as a versatile generalization that encompasses several distributions as special cases. Specifically, when the logarithmic Hamiltonian term vanishes, i.e., $K(t,x,i,V_x,V_{xx}) = \mathbf{0}$, the shape parameters become $\alpha_k = 1$, and the optimal exploratory policy $\boldsymbol{\pi}^*(a\mid t,x,i)$ degenerates into the product of independent one-dimensional exponential distributions discussed in Subsection \ref{subsection Exponential distribution}. On the present open action space, their zero modal locations are limiting boundary peaks; they are attained if zero actions are admitted. 

Furthermore, when $L(t,x,i,V_x,V_{xx}) = -\frac{\lambda}{2}\mathbf{1}$, and the logarithmic coefficient vector takes the specific form $K(t,x,i,V_x,V_{xx}) = \lambda\left(\frac{1}{2}\boldsymbol{\nu} - \mathbf{1}\right)$, where $\boldsymbol{\nu} = (\nu_1, \dots, \nu_n)^\top \in (\mathbb{R}_{>0})^n$ specifies the degrees of freedom and $\mathbf{1} = (1, \dots, 1)^\top \in \mathbb{R}^n$, the policy $\boldsymbol{\pi}^*(a\mid t,x,i)$ degenerates into the product of $n$ independent one-dimensional Chi-squared distributions with degrees of freedom $\nu_1, \nu_2, \dots, \nu_n$, respectively.
In this special case,
\begin{equation}
\mathbb{E}^{\boldsymbol{\pi}^*}[a_k]=\nu_k,\quad
\text{Var}^{\boldsymbol{\pi}^*}[a_k]=2\nu_k,\quad
a_k^{\mathrm{mode}}=\nu_k-2\quad(\nu_k>2).
\end{equation}
For $0<\nu_k\leq2$, the limiting boundary peak is at zero.
\end{remark}

\subsection{Beta distribution}\label{subsection Beta distribution}
Suppose the control space is the open unit hypercube, i.e., $\mathcal{A}=(0, 1)^n \subseteq \mathbb{R}^n$, which is used in applications where the control variables represent proportions, rates, or fractional allocations (e.g., portfolio weights). To generate a well-defined bounded exploratory distribution, we assume the Hamiltonian is additively separable with respect to the control components and incorporates logarithmic barrier terms for both boundaries $0$ and $1$. Specifically, the Hamiltonian can be expressed in vector form as
\begin{equation}
\begin{aligned}
H(t,x,i,V_x,V_{xx};a)
=&r(t, x, i;a)+b(t, x, i; a) \circ V_x + \frac{1}{2}\sigma^2(t, x, i; a) \circ V_{xx}\\
=&K(t,x,i,V_x,V_{xx})^\top \ln(a) + L(t,x,i,V_x,V_{xx})^\top \ln(\mathbf{1}-a)\\
&+ C(t,x,i,V_x,V_{xx}),
\end{aligned}
\end{equation}
where $K(t,x,i,V_x,V_{xx}) \in (-\lambda, \infty)^n$ and $L(t,x,i,V_x,V_{xx}) \in (-\lambda, \infty)^n$ are vector-valued functions, $C(t,x,i,V_x,V_{xx}) \in \mathbb{R}$ is a scalar function, and $\mathbf{1} = (1, \dots, 1)^\top \in \mathbb{R}^n$. Here, $\ln(a)$ and $\ln(\mathbf{1}-a)$ denote the element-wise natural logarithms of the vectors $a$ and $\mathbf{1}-a$, respectively. The conditions $K_k > -\lambda$ and $L_k > -\lambda$ for each component $k=1,\dots,n$ are required to ensure the integrability of the probability density near the boundaries $a_k \to 0^+$ and $a_k \to 1^-$, guaranteeing a well-defined distribution over $(0,1)^n$. For notational simplicity, we drop the arguments and denote these vectors simply as $K$ and $L$.

According to Theorem \ref{Theorem Optimal strategy}, the optimal strategy is
\begin{equation}
\boldsymbol{\pi}^*(a\mid t,x,i)=\frac{\displaystyle\exp \left( \frac{1}{\lambda} K^\top \ln(a) + \frac{1}{\lambda} L^\top \ln(\mathbf{1}-a)+\frac{1}{\lambda}C(t,x,i,V_x,V_{xx}) \right)}{\displaystyle\int_{(0, 1)^n} \exp \left( \frac{1}{\lambda} K^\top \ln(a) + \frac{1}{\lambda} L^\top \ln(\mathbf{1}-a)+\frac{1}{\lambda}C(t,x,i,V_x,V_{xx}) \right) da}.
\end{equation}
The constant term $\exp(C/\lambda)$ cancels out. By expanding the dot products into sums and applying the algebraic identity $\exp(a \ln x) = x^a$, the joint probability density factors into the product of $n$ independent one-dimensional functions
\begin{equation}
\boldsymbol{\pi}^*(a\mid t,x,i) \propto \prod_{k=1}^n a_k^{\frac{K_k}{\lambda}} (1-a_k)^{\frac{L_k}{\lambda}}.
\end{equation}
By matching this functional form with the probability density function of a standard Beta distribution $f(x; \alpha, \beta) \propto x^{\alpha-1} (1-x)^{\beta-1}$, we can deduce that under the optimal exploratory policy, each control component $a_k$ is independent and follows a Beta distribution
\begin{equation}
a_k \sim \mathrm{Beta}\left(\alpha_k, \beta_k\right), \quad k = 1, \dots, n,
\end{equation}
where the shape parameters $\alpha_k$ and $\beta_k$ are given by
\begin{equation}
\alpha_k = \frac{K_k(t,x,i,V_x,V_{xx})}{\lambda} + 1, \quad \beta_k = \frac{L_k(t,x,i,V_x,V_{xx})}{\lambda} + 1.
\end{equation}

The expected optimal control action and the variance for each component under the exploratory policy are given by
\begin{equation}
\begin{aligned}
\mathbb{E}^{\boldsymbol{\pi}^*}[a_k] &= \frac{\alpha_k}{\alpha_k + \beta_k} = \frac{K_k + \lambda}{K_k + L_k + 2\lambda},\\
\text{Var}^{\boldsymbol{\pi}^*}[a_k] &= \frac{\alpha_k \beta_k}{(\alpha_k + \beta_k)^2 (\alpha_k + \beta_k + 1)} = \frac{(K_k + \lambda)(L_k + \lambda) \lambda}{(K_k + L_k + 2\lambda)^2 (K_k + L_k + 3\lambda)}.
\end{aligned}
\end{equation}

For $K_k>0$ and $L_k>0$, the marginal log-density is strictly concave,
and its first-order condition gives the unique interior mode
\begin{equation}
a_k^{\mathrm{mode}}
=\frac{\alpha_k-1}{\alpha_k+\beta_k-2}
=\frac{K_k}{K_k+L_k}.
\end{equation}
For the remaining admissible parameter ranges, the modal locations are
\begin{equation}
\mathcal{B}_k=
\begin{cases}
\{0\}, & K_k\leq0,\ L_k\geq0,\ (K_k,L_k)\ne(0,0),\\
\{1\}, & K_k\geq0,\ L_k\leq0,\ (K_k,L_k)\ne(0,0),\\
\{0,1\}, & K_k<0,\ L_k<0,\\
(0,1), & K_k=L_k=0.
\end{cases}
\end{equation}
Here $0$ and $1$ denote limiting boundary peaks, since neither endpoint
belongs to the stated action space. When $K_k=L_k=0$, the density is
uniform and every admissible point is a mode.

\begin{remark}
For $K_k,L_k>0$, the Beta mode $K_k/(K_k+L_k)$ has the same functional form as the classical deterministic allocation obtained by maximizing the Hamiltonian. At a positive temperature, the mean $(K_k+\lambda)/(K_k+L_k+2\lambda)$ generally differs from this mode; they coincide in the symmetric case $K_k=L_k$. With positive $K_k$ and $L_k$ fixed, increasing $\lambda$ moves the mean towards $1/2$ while leaving the mode unchanged, and the density approaches the uniform distribution as $\lambda\to\infty$. As $\lambda\to0^+$, the variance vanishes and the distribution concentrates at its mode. 
\end{remark}

\subsection{Dirichlet distribution}
Suppose the control space is the open $(n-1)$-simplex, i.e., $\mathcal{A} = \{ a \in (0, 1)^n \mid \mathbf{1}^\top a = 1 \}$, where $n\geq2$ and $\mathbf{1} = (1, \dots, 1)^\top \in \mathbb{R}^n$. Densities and integrals on this simplex are understood with respect to the coordinates $(a_1,\dots,a_{n-1})$, with $a_n=1-\sum_{k=1}^{n-1}a_k$. This action space is ubiquitous in resource allocation and portfolio optimization problems where the control variables represent fractional weights that must sum to unity. Specifically, we assume the Hamiltonian contains only logarithmic terms of the control components
\begin{equation}
\begin{aligned}
H(t,x,i,V_x,V_{xx};a)
=&r(t, x, i;a)+b(t, x, i; a) \circ V_x + \frac{1}{2}\sigma^2(t, x, i; a) \circ V_{xx}\\
=&K(t,x,i,V_x,V_{xx})^\top \ln(a) + C(t,x,i,V_x,V_{xx}),
\end{aligned}
\end{equation}
where $K(t,x,i,V_x,V_{xx}) \in (-\lambda, \infty)^n$ is a vector-valued function, $C(t,x,i,V_x,V_{xx}) \in \mathbb{R}$ is a scalar function. The condition $K_k > -\lambda$ for all $k=1,\dots,n$ ensures that the probability density is integrable near the boundaries of the simplex. 

According to Theorem \ref{Theorem Optimal strategy}, the optimal strategy is
\begin{equation}
\begin{aligned}
\boldsymbol{\pi}^*(a\mid t,x,i)=&\frac{\displaystyle\exp \left( \frac{1}{\lambda} K(t,x,i,V_x,V_{xx})^\top \ln(a) +\frac{1}{\lambda}C(t,x,i,V_x,V_{xx})\right)}{\displaystyle\int_{\mathcal{A}} \exp \left( \frac{1}{\lambda} K(t,x,i,V_x,V_{xx})^\top \ln(a) +\frac{1}{\lambda}C(t,x,i,V_x,V_{xx})\right) da}\\
\propto&\prod_{k=1}^n a_k^{\frac{K_k}{\lambda}}.
\end{aligned}
\end{equation}
By matching this functional form with the probability density function of a Dirichlet distribution $f(x; \alpha) \propto \prod_{k=1}^n x_k^{\alpha_k - 1}$, we deduce that the optimal exploratory policy follows a Dirichlet distribution
\begin{equation}
a \sim \mathrm{Dirichlet}(\alpha_1, \dots, \alpha_n),
\end{equation}
where the concentration parameters are given by $\alpha_k = \frac{K_k}{\lambda} + 1$. 

Let $S=\sum_{j=1}^n K_j$ and $\alpha_0=\sum_{j=1}^n\alpha_j=S/\lambda+n$.
The mean, variance, and covariance are given by
\begin{equation}
\begin{aligned}
\mathbb{E}^{\boldsymbol{\pi}^*}[a_k]
&=\frac{\alpha_k}{\alpha_0}=\frac{K_k+\lambda}{S+n\lambda},\\
\text{Var}^{\boldsymbol{\pi}^*}[a_k]
&=\frac{\alpha_k(\alpha_0-\alpha_k)}{\alpha_0^2(\alpha_0+1)}
=\frac{\lambda(K_k+\lambda)(S-K_k+(n-1)\lambda)}
{(S+n\lambda)^2(S+(n+1)\lambda)},\\
\text{Cov}^{\boldsymbol{\pi}^*}[a_k,a_\ell]
&=-\frac{\alpha_k\alpha_\ell}{\alpha_0^2(\alpha_0+1)}
=-\frac{\lambda(K_k+\lambda)(K_\ell+\lambda)}
{(S+n\lambda)^2(S+(n+1)\lambda)},\quad k\ne\ell.
\end{aligned}
\end{equation}
If every $K_k>0$, maximizing $\sum_k K_k\ln a_k$ subject to
$\sum_k a_k=1$ gives $K_k/a_k=\eta$ for a common Lagrange multiplier
$\eta=S$. Thus the unique joint mode has coordinates
\begin{equation}
a_k^{\mathrm{mode}}
=\frac{\alpha_k-1}{\alpha_0-n}
=\frac{K_k}{S},\qquad k=1,\dots,n.
\end{equation}
These are coordinates of the joint mode. For comparison, the marginal law
is $\mathrm{Beta}(\alpha_k,\alpha_0-\alpha_k)$, whose mode is
$(\alpha_k-1)/(\alpha_0-2)$ when both marginal shape parameters exceed one.
If all $K_k\geq0$, $S>0$, and some $K_k=0$, the vector $(K_k/S)_{k=1}^n$
is instead the unique maximizer of the continuously extended density on
the closed simplex and is not attained in the open simplex. If all
$K_k=0$, the distribution is uniform and every point of the open simplex
is a mode. If some $K_k<0$, the density and Hamiltonian are unbounded
near the corresponding boundary face, and no finite maximizer exists
in the open simplex.

\begin{remark}
The Dirichlet distribution respects the constraint $\sum_k a_k=1$. When all $K_k>0$, its joint mode $K_k/S$ has the same functional form as the classical deterministic resource allocation. The mean $(K_k+\lambda)/(S+n\lambda)$ generally differs from the mode at a positive temperature, although the two vectors coincide when all $K_k$ are equal. With the positive coefficients fixed, the covariance matrix vanishes and the policy concentrates at the modal allocation as $\lambda\to0^+$. As $\lambda\to\infty$, the density approaches the uniform distribution on the simplex and its mean approaches $(1/n,\dots,1/n)$. 
\end{remark}

\subsection{Weibull distribution}
Suppose the control space is the positive orthant, i.e., $\mathcal{A}=(\mathbb{R}_{> 0})^n \subseteq \mathbb{R}^n$, and the system incorporates non-linear, power-law penalties on the control effort (e.g., non-linear aerodynamic drag, or fractional-order costs). To capture this, we assume the Hamiltonian contains both logarithmic terms and arbitrary power terms of the control components. Specifically, it can be expressed in vector form as
\begin{equation}
\begin{aligned}
H(t,x,i,V_x,V_{xx};a)
=&r(t, x, i;a)+b(t, x, i; a) \circ V_x + \frac{1}{2}\sigma^2(t, x, i; a) \circ V_{xx}\\
=&\lambda(p - \mathbf{1})^\top \ln(a) + L(t,x,i,V_x,V_{xx})^\top a^p \\
&+ C(t,x,i,V_x,V_{xx}),
\end{aligned}
\end{equation}
where $L(t,x,i,V_x,V_{xx}) \in (\mathbb{R}_{<0})^n$ is a vector-valued function, $C(t,x,i,V_x,V_{xx}) \in \mathbb{R}$ is a scalar function, and $p = (p_1, \dots, p_n)^\top \in (\mathbb{R}_{>0})^n$ dictates the polynomial degree of the cost for each component. Here, $\ln(a) := (\ln a_1, \dots, \ln a_n)^\top$ denotes the element-wise natural logarithm, and $a^p := (a_1^{p_1}, \dots, a_n^{p_n})^\top$ denotes the element-wise power operation. The negative constraint $L_k < 0$ guarantees exponential tail decay as $a_k \to \infty$. 

According to Theorem \ref{Theorem Optimal strategy}, the optimal strategy is
\begin{equation}
\begin{aligned}
\boldsymbol{\pi}^*(a\mid t,x,i)&=\frac{\displaystyle\exp \left( \frac{1}{\lambda} \big(\lambda(p-\mathbf{1})\big)^\top \ln(a) + \frac{1}{\lambda} L(t,x,i,V_x,V_{xx})^\top a^p \right)}{\displaystyle\int_{(\mathbb{R}_{> 0})^n} \exp \left( \frac{1}{\lambda} \big(\lambda(p-\mathbf{1})\big)^\top \ln(a) + \frac{1}{\lambda} L(t,x,i,V_x,V_{xx})^\top a^p \right) da} \\
&\propto \prod_{k=1}^n a_k^{p_k - 1} \exp\left( \frac{L_k(t,x,i,V_x,V_{xx})}{\lambda} a_k^{p_k} \right),
\end{aligned}
\end{equation}
where $\mathbf{1} = (1, \dots, 1)^\top \in \mathbb{R}^n$.
By matching this form with the probability density function of the standard Weibull distribution $f(x; \alpha, \sigma) = \frac{\alpha}{\sigma} \left(\frac{x}{\sigma}\right)^{\alpha-1} \exp\left(-\left(\frac{x}{\sigma}\right)^\alpha\right)$, we can deduce that under the optimal exploratory policy, each control component $a_k$ is independent and follows a Weibull distribution
\begin{equation}
a_k \sim \mathrm{Weibull}\left(\alpha_k, \sigma_k\right), \quad k = 1, \dots, n,
\end{equation}
where the shape parameter $\alpha_k$ and the scale parameter $\sigma_k$ are given by
\begin{equation}
\alpha_k = p_k, \quad \sigma_k = \left( -\frac{\lambda}{L_k(t,x,i,V_x,V_{xx})} \right)^{\frac{1}{p_k}}.
\end{equation}

The expected optimal control action and the variance under this Weibull exploratory policy are given by
\begin{equation}
\begin{aligned}
\mathbb{E}^{\boldsymbol{\pi}^*}[a_k] &= \sigma_k \Gamma\left(1 + \frac{1}{\alpha_k}\right) = \left( -\frac{\lambda}{L_k} \right)^{\frac{1}{p_k}} \Gamma\left(1 + \frac{1}{p_k}\right),\\
\text{Var}^{\boldsymbol{\pi}^*}[a_k] &= \sigma_k^2 \left[ \Gamma\left(1 + \frac{2}{\alpha_k}\right) - \left( \Gamma\left(1 + \frac{1}{\alpha_k}\right) \right)^2 \right] \\
&= \left( -\frac{\lambda}{L_k} \right)^{\frac{2}{p_k}} \left[ \Gamma\left(1 + \frac{2}{p_k}\right) - \Gamma^2\left(1 + \frac{1}{p_k}\right) \right],
\end{aligned}
\end{equation}
where $\Gamma(\cdot)$ is the standard Gamma function.

The logarithmic derivative of the marginal density is
\begin{equation}
\frac{d}{da_k}\ln\pi_k^*(a_k)
=\frac{p_k-1}{a_k}+\frac{p_kL_k}{\lambda}a_k^{p_k-1}.
\end{equation}
Its modal location is therefore
\begin{equation}
a_k^{\mathrm{mode}}=
\begin{cases}
\displaystyle\sigma_k\left(\frac{p_k-1}{p_k}\right)^{1/p_k}
=\left(-\frac{\lambda(p_k-1)}{p_kL_k}\right)^{1/p_k},
& p_k>1,\\[6pt]
0\quad\text{(limiting boundary peak)},&0<p_k\leq1.
\end{cases}
\end{equation}
For $p_k>1$, this is the unique interior maximum. For $0<p_k\leq1$,
the density is strictly decreasing, so there is no attained mode on
$(0,\infty)$. If $p_k<1$, the Hamiltonian is unbounded above near zero;
if $p_k=1$, it has a finite, unattained boundary supremum.

\subsection{Pareto distribution}
Suppose the control space is bounded from below by a positive threshold, i.e., $\mathcal{A}=[m, \infty) := [m_1, \infty) \times \dots \times [m_n, \infty) \subset \mathbb{R}^n$, where $m = (m_1, \dots, m_n)^\top \in (\mathbb{R}_{> 0})^n$ represents the minimum necessary control effort or baseline activation threshold for each component. To capture a scale-free or power-law decay in exploration, we assume the Hamiltonian imposes a pure logarithmic penalty on the control effort, with no polynomial or linear terms. Specifically, it can be expressed in vector form as
\begin{equation}
\begin{aligned}
H(t,x,i,V_x,V_{xx};a)
=&r(t, x, i;a)+b(t, x, i; a) \circ V_x + \frac{1}{2}\sigma^2(t, x, i; a) \circ V_{xx}\\
=&K(t,x,i,V_x,V_{xx})^\top \ln(a) + C(t,x,i,V_x,V_{xx}),
\end{aligned}
\end{equation}
where $K(t,x,i,V_x,V_{xx}) \in (\mathbb{R}_{<0})^n$ is a vector-valued function, and $C(t,x,i,V_x,V_{xx}) \in \mathbb{R}$ is a scalar function. Here, $\ln(a)$ denotes the element-wise natural logarithm. To ensure the probability density is normalizable over the semi-infinite domain $[m, \infty)$, the logarithmic penalty must be sufficiently steep; specifically, we require $K_k < -\lambda$ for all $k = 1, \dots, n$.

According to Theorem \ref{Theorem Optimal strategy}, the optimal strategy is
\begin{equation}
\begin{aligned}
\boldsymbol{\pi}^*(a\mid t,x,i)&=\frac{\displaystyle\exp \left( \frac{1}{\lambda} K(t,x,i,V_x,V_{xx})^\top \ln(a) \right)}{\displaystyle\int_{[m, \infty)} \exp \left( \frac{1}{\lambda} K(t,x,i,V_x,V_{xx})^\top \ln(a) \right) da} \\
&\propto \prod_{k=1}^n a_k^{\frac{K_k}{\lambda}},\quad a_k \ge m_k.
\end{aligned}
\end{equation}
By matching this functional form with the probability density function of the Pareto distribution $f(x; m, \alpha) \propto x^{-(\alpha+1)}$ (for $x \ge m$), we can deduce that under the optimal exploratory policy, each control component $a_k$ follows a Pareto distribution
\begin{equation}
a_k \sim \mathrm{Pareto}\left(m_k, \alpha_k\right), \quad k = 1, \dots, n,
\end{equation}
where the scale parameter is strictly determined by the domain boundary $m_k$, and the shape parameter $\alpha_k$ is given by
\begin{equation}
\alpha_k = -\frac{K_k(t,x,i,V_x,V_{xx})}{\lambda} - 1.
\end{equation}
Notice that the integrability condition $K_k < -\lambda$ guarantees that $\alpha_k > 0$, ensuring a mathematically valid Pareto distribution.

The expected optimal control action and the variance are given by
\begin{equation}
\begin{aligned}
\mathbb{E}^{\boldsymbol{\pi}^*}[a_k] &= \frac{\alpha_k m_k}{\alpha_k - 1} = \frac{(K_k + \lambda)m_k}{K_k + 2\lambda}, \quad \text{for } K_k < -2\lambda \ (\alpha_k > 1),\\
\text{Var}^{\boldsymbol{\pi}^*}[a_k] &= \frac{m_k^2 \alpha_k}{(\alpha_k - 1)^2 (\alpha_k - 2)} = \frac{m_k^2 \lambda^2(K_k + \lambda)}{(K_k + 2\lambda)^2 (K_k + 3\lambda)}, \quad \text{for } K_k < -3\lambda \ (\alpha_k > 2).
\end{aligned}
\end{equation}

Since $K_k<0$, the marginal density is strictly decreasing on
$[m_k,\infty)$, and its unique mode is
\begin{equation}
a_k^{\mathrm{mode}}=m_k,\qquad k=1,\dots,n.
\end{equation}
Thus the joint mode is $a^{\mathrm{mode}}=m$ for every normalizable
policy, including cases where the mean or variance is not finite.

\subsection{Log-Normal distribution}
Suppose the control space is the positive orthant, i.e., $\mathcal{A}=(\mathbb{R}_{> 0})^n \subseteq \mathbb{R}^n$, which is common in multiplicative systems such as geometric Brownian motion models in finance or population dynamics. We assume the Hamiltonian is quadratic with respect to the logarithm of the control variable. Specifically, to match the density of a log-normal distribution, the Hamiltonian can be expressed as
\begin{equation}
\begin{aligned}
H(t,x,i,V_x,V_{xx};a)
=&r(t, x, i;a)+b(t, x, i; a) \circ V_x + \frac{1}{2}\sigma^2(t, x, i; a) \circ V_{xx}\\
=&-\frac{1}{2} \ln(a)^\top R(t,x,i,V_x,V_{xx}) \ln(a) + L(t,x,i,V_x,V_{xx})^\top \ln(a) \\
&- \lambda \mathbf{1}^\top \ln(a) + C(t,x,i,V_x,V_{xx}),
\end{aligned}
\end{equation}
where $\ln(a) := (\ln a_1, \dots, \ln a_n)^\top$, $R(t,x,i,V_x,V_{xx}) \in \mathbb{R}^{n \times n}$ is a symmetric positive definite matrix, $L(t,x,i,V_x,V_{xx}) \in \mathbb{R}^n$ is a vector, $C(t,x,i,V_x,V_{xx}) \in \mathbb{R}$ is a scalar, and $\mathbf{1} = (1, \dots, 1)^\top \in \mathbb{R}^n$. The positive definite nature of $R$ ensures that the quadratic term dominates as $\ln(a) \to \pm\infty$, guaranteeing the distribution is normalizable over $(\mathbb{R}_{> 0})^n$. For notational simplicity, we drop the arguments and denote these terms simply as $R$, $L$, and $C$.

According to Theorem \ref{Theorem Optimal strategy}, the optimal strategy is
\begin{equation}
\begin{aligned}
\boldsymbol{\pi}^*(a\mid t,x,i)&=\frac{\displaystyle\exp \left( \frac{1}{\lambda} \left( -\frac{1}{2} \ln(a)^\top R \ln(a) + L^\top \ln(a) - \lambda \mathbf{1}^\top \ln(a) \right) \right)}{\displaystyle\int_{(\mathbb{R}_{> 0})^n} \exp \left( \frac{1}{\lambda} \left( -\frac{1}{2} \ln(a)^\top R \ln(a) + L^\top \ln(a) - \lambda \mathbf{1}^\top \ln(a) \right) \right) da}.
\end{aligned}
\end{equation}
By isolating the $-\lambda \mathbf{1}^\top \ln(a)$ term, which translates to $\prod_{k=1}^n a_k^{-1}$, and completing the square for the remaining terms involving $\ln(a)$, the probability density simplifies to
\begin{equation}
\boldsymbol{\pi}^*(a\mid t,x,i) \propto\left( \prod_{k=1}^n \frac{1}{a_k} \right) \exp \left( -\frac{1}{2} \left( \ln(a) - R^{-1}L \right)^\top \left( \frac{ R}{\lambda} \right) \left( \ln(a) - R^{-1}L \right) \right).
\end{equation}
This is the functional form of a multivariate Log-Normal distribution. Thus, under the optimal exploratory policy, the control vector follows
\begin{equation}
a \sim \mathrm{LogNormal}\left( R(t,x,i,V_x,V_{xx})^{-1} L(t,x,i,V_x,V_{xx}), \lambda R(t,x,i,V_x,V_{xx})^{-1} \right).
\end{equation}
Write $\mu=R^{-1}L$ and $\Sigma=\lambda R^{-1}$, so that
$\ln a\sim\mathcal{N}(\mu,\Sigma)$. The mean, variance, and covariance are
\begin{equation}
\begin{aligned}
\mathbb{E}^{\boldsymbol{\pi}^*}[a_k]
&=\exp\left(\mu_k+\frac12\Sigma_{kk}\right),\\
\text{Var}^{\boldsymbol{\pi}^*}[a_k]
&=\left(e^{\Sigma_{kk}}-1\right)e^{2\mu_k+\Sigma_{kk}},\\
\text{Cov}^{\boldsymbol{\pi}^*}[a_k,a_\ell]
&=\left(e^{\Sigma_{k\ell}}-1\right)
\exp\left(\mu_k+\mu_\ell+\frac12(\Sigma_{kk}+\Sigma_{\ell\ell})\right).
\end{aligned}
\end{equation}
To find the joint mode with respect to the action coordinates $a$, let
$y=\ln a$ and maximize the original log-density evaluated at $a=e^y$:
\begin{equation}
\ln\boldsymbol{\pi}^*(e^y\mid t,x,i)
=-\mathbf{1}^{\top}y-\frac12(y-\mu)^{\top}\Sigma^{-1}(y-\mu)
+\text{constant}.
\end{equation}
This function is strictly concave in $y$. Its first-order condition
$-\mathbf{1}-\Sigma^{-1}(y-\mu)=0$ gives the unique joint mode
\begin{equation}\label{eq:lognormal_joint_mode}
a^{\mathrm{mode}}
=\exp\left(\mu-\Sigma\mathbf{1}\right)
=\exp\left(R^{-1}(L-\lambda\mathbf{1})\right),
\end{equation}
where the exponential is applied componentwise. 

\section{Structure-informed $q$-learning}\label{Section 4}
Section~\ref{Section 3} identifies explicit policy distributions from the
action domain and the dependence of the Hamiltonian on the action. We now
use this information to construct a learning algorithm. The distributional
family, its support, and any known parameter relations are specified by
the control problem, while the remaining parameter functions are learned
from observed trajectories. In this sense, the results of
Section~\ref{Section 3} provide a \emph{distributional structure prior}.
This term refers to information used in parameterization; no Bayesian
prior or additional relative-entropy penalty is introduced.

We assume that the state, the regime, the implemented action, and the
reward are observable. The numerical coefficients of $b$ and $\sigma$,
and the transition rates $Q$, need not be supplied to the learner.
However, the action dependence needed to select a distributional family
is assumed known. Thus the algorithm is model-free with respect to these
unknown coefficients while retaining structural information about the
control problem. Our construction builds on the continuous-time
$q$-learning framework of \citet{jia2023q}, \citet{zhang2026continuous} and uses the discrete action
sampling formulation in \citet{jia2025erratum}.

\subsection{$q$-functions and structure-preserving policy improvement}
\label{sub41}
Based on the structure of the exploratory HJB equation \eqref{HJB equation}, we introduce the $q$-function, which evaluates the action-value of a deterministic action $a \in \mathcal{A}$ before the agent acts according to its stochastic policy.

\begin{definition}\label{definition q-function}
($q$-Function). For a given policy $\boldsymbol{\pi}(\cdot \mid t, x, i)\in\boldsymbol{\Pi}$, let $J(t,x,i; \boldsymbol{\pi})$ be the corresponding exploratory payoff function. The $q$-function is defined as
\begin{equation}\label{q_function}
\begin{aligned}
q(t, x, i; a; \boldsymbol{\pi}) :=&  J_t(t, x, i; \boldsymbol{\pi}) + H(t, x, i, J_x,  J_{xx}; a) \\
&+ \sum_{j=1}^m q_{ij} J(t, x, j; \boldsymbol{\pi}).
\end{aligned}
\end{equation}
\end{definition}

It immediately follows from the exploratory HJB equation that the $q$-function satisfies the following consistency condition,
\begin{equation}\label{q_consistency}
\int_{\mathcal{A}} \left( q(t, x, i; a; \boldsymbol{\pi}) - \lambda \ln \boldsymbol{\pi}(a \mid t, x, i) \right) \boldsymbol{\pi}(a \mid t, x, i) \,da = 0, \quad \forall(t, x, i).
\end{equation}

\begin{lemma}\label{Policy improvement}
(Policy improvement)
Let $\boldsymbol{\pi}(a\mid t,x,i)\in \boldsymbol{\Pi}$ be an arbitrarily given admissible policy. Suppose that the corresponding exploratory payoff function $J(\cdot,\cdot,i;\boldsymbol{\pi}) \in C^{1,2}([0,T) \times \mathbb{R}^d) \cap C^{0}([0,T] \times \mathbb{R}^d)$ for each $i \in \mathcal{S}$. For any $(t,x,i)\in[0,T) \times \mathbb{R}^d \times \mathcal{S}$, the policy $\tilde{\boldsymbol{\pi}}(a\mid t,x,i)$ defined by
\begin{equation}\label{Policy improvement pi}
\tilde{\boldsymbol{\pi}}(a\mid t,x,i)=\frac{\displaystyle\exp \left(\frac{1}{\lambda}H(t,x,i,J_x,J_{xx};a)\right)}{\displaystyle\int_\mathcal{A} \exp \left(\frac{1}{\lambda}H(t,x,i,J_x,J_{xx};a)\right)da} 
\end{equation}
is admissible. Then, the exploratory payoff function under the specific policies $\tilde{\boldsymbol{\pi}}(a\mid t,x,i)$ and $\boldsymbol{\pi}(a\mid t,x,i)$ satisfies
\begin{equation}
J(t,x,i;\tilde{\boldsymbol{\pi}})\geq J(t,x,i;\boldsymbol{\pi}),\quad \forall (t,x,i)\in[0,T]\times\mathbb{R}^d \times \mathcal{S}.
\end{equation}
\end{lemma}
\begin{proof}
Fix $(t, x, i)\in [0,T]\times\mathbb{R}^d \times \mathcal{S}$. Let $X_h^{\tilde{\boldsymbol{\pi}}}$ be the state process driven by the updated policy $\tilde{\boldsymbol{\pi}}(a\mid t,x,i)$. By applying It\^{o}'s formula with regime switching to $J(t,x,i;\boldsymbol{\pi})$ along the path of $X_h^{\tilde{\boldsymbol{\pi}}}$, we have
\begin{equation*}
\begin{aligned}
&J(s,X_{s}^{\tilde{\boldsymbol{\pi}}}, \theta_s;\boldsymbol{\pi})-J(t,x, i;\boldsymbol{\pi})\\
=&\int^s_tJ_x(h, X_{h-}^{\tilde{\boldsymbol{\pi}}}, \theta_{h-};\boldsymbol{\pi})\circ\tilde{\sigma}(h,X_{h-}^{\tilde{\boldsymbol{\pi}}}, \theta_{h-};\tilde{\boldsymbol{\pi}}_h)dW_h+\int_t^s dM^{\theta}_h \\
&+\int^s_{t}\Big[\tilde{b}(h,X_{h-}^{\tilde{\boldsymbol{\pi}}}, \theta_{h-};\tilde{\boldsymbol{\pi}}_h) \circ J_x(h,X_{h-}^{\tilde{\boldsymbol{\pi}}}, \theta_{h-};\boldsymbol{\pi})\\
&\quad+\frac{1}{2} \tilde{\sigma}^2(h,X_{h-}^{\tilde{\boldsymbol{\pi}}}, \theta_{h-};\tilde{\boldsymbol{\pi}}_h) \circ J_{xx}(h,X_{h-}^{\tilde{\boldsymbol{\pi}}}, \theta_{h-};\boldsymbol{\pi})\\
&\quad+J_t(h,X_{h-}^{\tilde{\boldsymbol{\pi}}}, \theta_{h-};\boldsymbol{\pi})+\sum_{j=1}^m q_{\theta_{h-}, j} J(h, X_{h-}^{\tilde{\boldsymbol{\pi}}}, j;\boldsymbol{\pi})\Big]dh,
\end{aligned}
\end{equation*}
where $M^{\theta}_h$ is the zero-mean local martingale associated with the jumps of the continuous-time Markov chain $\theta$.
Define the stopping times
\begin{equation}
\tau_n := \inf\left\{s \geq t : \int_t^s \big|J_x(h, X_{h-}^{\tilde{\boldsymbol{\pi}}}, \theta_{h-};\boldsymbol{\pi})\circ \tilde{\sigma}(h,X_{h-}^{\tilde{\boldsymbol{\pi}}}, \theta_{h-};\tilde{\boldsymbol{\pi}}_h)\big|^2dh \geq n \right\}, \,\text{for}\,\, n \geq 1.
\end{equation} By taking the conditional expectation, the martingale terms vanish, and we obtain
\begin{equation}\label{Ja}
\begin{aligned}	
&J(s\wedge\tau_n,X_{s\wedge\tau_n}^{\tilde{\boldsymbol{\pi}}}, \theta_{s\wedge\tau_n};\boldsymbol{\pi})-J(t,x,i;\boldsymbol{\pi})\\
=&\E_{t,x,i}\left[ \int^{s\wedge\tau_n}_{t}\Big(\tilde{b}(h,X_{h-}^{\tilde{\boldsymbol{\pi}}}, \theta_{h-};\tilde{\boldsymbol{\pi}}_h) \circ J_x(h, X_{h-}^{\tilde{\boldsymbol{\pi}}}, \theta_{h-};\boldsymbol{\pi})\right.\\
&\quad+\frac{1}{2} \tilde{\sigma}^2(h,X_{h-}^{\tilde{\boldsymbol{\pi}}}, \theta_{h-};\tilde{\boldsymbol{\pi}}_h) \circ J_{xx}(h, X_{h-}^{\tilde{\boldsymbol{\pi}}}, \theta_{h-};\boldsymbol{\pi})\\
&\quad\left.+J_t(h, X_{h-}^{\tilde{\boldsymbol{\pi}}}, \theta_{h-};\boldsymbol{\pi})+\sum_{j=1}^m q_{\theta_{h-}, j} J(h, X_{h-}^{\tilde{\boldsymbol{\pi}}}, j;\boldsymbol{\pi})\Big)dh\right].
\end{aligned}
\end{equation}

On the other hand, based on the HJB equation for the current policy $\boldsymbol{\pi}(a\mid t,x,i)$ and the definition of the optimal response $\tilde{\boldsymbol{\pi}}(a\mid t,x,i)$, we have the following inequality $\forall(t,x,i)\in[0, T) \times \mathbb{R}^d \times \mathcal{S}$:
\begin{equation}\label{Ha}
\begin{aligned}
-J_t(t,x,i;\boldsymbol{\pi})
=&\int_{\mathcal{A}}H(t, x, i, J_x, J_{xx};a)\,\boldsymbol{\pi}(a \mid t, x, i)\,da \\
&+\sum_{j=1}^{m} q_{ij} J(t, x, j;\boldsymbol{\pi})-\lambda\int_{\mathcal{A}}\boldsymbol{\pi}(a \mid t, x, i) \ln \boldsymbol{\pi}(a \mid t, x, i)\,da\\
\leq&\int_{\mathcal{A}}H(t, x, i, J_x, J_{xx};a)\,\tilde{\boldsymbol{\pi}}(a \mid t, x, i)\,da \\
&+\sum_{j=1}^{m} q_{ij} J(t, x, j;\boldsymbol{\pi})-\lambda\int_{\mathcal{A}}\tilde{\boldsymbol{\pi}}(a \mid t, x, i) \ln \tilde{\boldsymbol{\pi}}(a \mid t, x, i)\,da.
\end{aligned}
\end{equation}

Substituting \eqref{Ha} into \eqref{Ja} yields
$$\begin{aligned}
J(t,x,i;\boldsymbol{\pi})
\leq& \E_{t,x,i}\bigg[\int_t^{s\wedge\tau_n} \Big(\tilde{r}(h,X_{h-}^{\tilde{\boldsymbol{\pi}}}, \theta_{h-};\tilde{\boldsymbol{\pi}}_h)\\
&-\lambda\int_{\mathcal{A}} \tilde{\boldsymbol{\pi}}_h(a \mid t, x, i) \ln \tilde{\boldsymbol{\pi}}_h(a \mid t, x, i)\, da \Big)d h+J(s\wedge\tau_n,X^{\tilde{\boldsymbol{\pi}}}_{s\wedge\tau_n}, \theta_{s\wedge\tau_n};\boldsymbol{\pi})\bigg].
\end{aligned}$$

Set $s = T$ and let $n \to \infty$. By applying the dominated convergence theorem and combining with the terminal condition $J(T, X_T^{\tilde{\boldsymbol{\pi}}}, \theta_T;\boldsymbol{\pi})= g(X_T^{\tilde{\boldsymbol{\pi}}}, \theta_T)$, we derive that $\forall(t,x,i) \in [0,T] \times \mathbb{R}^d \times \mathcal{S}$,
$$\begin{aligned}
J(t,x,i;\boldsymbol{\pi})
\leq& \E_{t,x,i}\bigg[\int_t^{T} \Big(\tilde{r}(h,X_{h-}^{\tilde{\boldsymbol{\pi}}}, \theta_{h-};\tilde{\boldsymbol{\pi}}_h)-\lambda\int_{\mathcal{A}} \tilde{\boldsymbol{\pi}}_h(a \mid t, x, i) \ln \tilde{\boldsymbol{\pi}}_h(a \mid t, x, i)\, da \Big)d h \\
&+ g(X_T^{\tilde{\boldsymbol{\pi}}}, \theta_T)\bigg]
=J(t,x,i;\tilde{\boldsymbol{\pi}}).
\end{aligned}$$
\end{proof}

\begin{theorem}\label{Policy improvement via Q-function}
(Policy improvement via $q$-function).
Let $\boldsymbol{\pi}(\cdot \mid t, x, i) \in \boldsymbol{\Pi}$ be an arbitrarily given admissible policy, with the corresponding exploratory payoff function $J(t, x, i; \boldsymbol{\pi})$ and $q$-function $q(t, x, i; a; \boldsymbol{\pi})$. Suppose that $J(\cdot,\cdot,i; \boldsymbol{\pi}) \in C^{1,2}([0,T) \times \mathbb{R}^d) \cap C^{0}([0,T] \times \mathbb{R}^d)$ for each $i \in \mathcal{S}$. For any $(t,x,i) \in [0,T) \times \mathbb{R}^d \times \mathcal{S}$, if the updated strategy $\tilde{\boldsymbol{\pi}}(a \mid t,x,i)$ defined by
\begin{equation}\label{q_str_improvement}
\tilde{\boldsymbol{\pi}}(a \mid t,x,i) = \frac{\displaystyle\exp \left( \frac{1}{\lambda} q(t, x, i; a; \boldsymbol{\pi}) \right)}{\displaystyle\int_{\mathcal{A}} \exp \left( \frac{1}{\lambda} q(t, x, i; a'; \boldsymbol{\pi}) \right) da'}
\end{equation}
is admissible, then the exploratory payoff function under the new strategy satisfies
\begin{equation}
J(t,x,i; \tilde{\boldsymbol{\pi}}) \geq J(t,x,i; \boldsymbol{\pi}).
\end{equation}
\end{theorem}
\begin{proof}
The proof follows directly from Definition \ref{definition q-function} and Lemma \ref{Policy improvement}.
\end{proof}

The HJB equation and the Gibbs formula yield
\begin{equation}\label{eq:sec4_optimal_normalization}
q^*(t,x,i;a)=\lambda\ln\boldsymbol{\pi}^*(a\mid t,x,i),
\quad
\int_{\mathcal A}\exp(q^*(t,x,i;a)/\lambda)\,da=1.
\end{equation}

\subsection{Distribution-informed parameterization}\label{sub4_distribution}
Fix $\lambda>0$. Let $\mathcal D$ be the family selected from
Section~\ref{Section 3} and let $f_{\mathcal D}(a;\eta)$ be its normalized
density. We use a value approximator $V_\phi(t,x,i)$ and a parameter map
\begin{equation}\label{eq:sec4_parameter_map}
\eta_\omega(t,x,i)
=\mathcal T_{\mathcal D}
 \bigl(\operatorname{NN}_\omega(t,x,i);\lambda\bigr).
\end{equation}
The map $\mathcal T_{\mathcal D}$ imposes parameter constraints and inserts
any quantities already known from the model. A shared network with a
regime input or separate output heads for the regimes can be used.
The policy and the candidate optimal $q$-function are defined together:
\begin{equation}\label{eq:sec4_shared_parameterization}
\begin{aligned}
\boldsymbol{\pi}_\omega(a\mid t,x,i)
&=f_{\mathcal D}(a;\eta_\omega(t,x,i)),\\
q_\omega(t,x,i;a)
&=\lambda\ln f_{\mathcal D}(a;\eta_\omega(t,x,i)).
\end{aligned}
\end{equation}
The normalization constraint in
\eqref{eq:sec4_optimal_normalization} then holds for every $\omega$.
The value network must also satisfy $V_\phi(T,x,i)=g(x,i)$.
When $g$ has the required smoothness, one possible construction is
$V_\phi(t,x,i)=g(x,i)+(T-t)v_\phi(t,x,i)$.

Table~\ref{tab:sec4_distribution_heads} summarizes the distributional
outputs. The product densities in that table are used for the
additively separable cases of Section~\ref{Section 3}; independence is
not imposed on a general coupled Hamiltonian. Only the unknown entries
are learned. For a positive scalar
parameter, an implementation can use
$\varepsilon+\operatorname{softplus}(u)$, where
$\operatorname{softplus}(u)=\ln(1+e^u)$ and $\varepsilon>0$ is a small
numerical lower bound. For a covariance matrix, use $BB^\top$ with a
lower triangular $B$ and positive diagonal entries. Such numerical
bounds restrict the approximation class and should be chosen consistently
with the parameter range of the problem.

\begin{table}[h!]
\caption{Distributional information used by the parameter map. In vector
cases, scalar positivity constraints apply componentwise.}
\label{tab:sec4_distribution_heads}
\centering
\small
\begin{tabular}{@{}p{0.15\textwidth}p{0.35\textwidth}p{0.42\textwidth}@{}}
\toprule
Family & Distribution parameters & Structural and parameter constraints\\
\midrule
Gaussian & Mean $\mu$ and covariance $\Sigma=BB^\top$
 & $B$ has a positive diagonal; correlations can be learned.\\[3pt]
Exponential & Rate vector $\beta$
 & $\beta_k>0$; the shape is fixed.\\[3pt]
Gamma & Shape $\alpha$ and rate $\beta$
 & $\alpha_k,\beta_k>0$; known shape or rate functions are fixed.\\[3pt]
Beta & Shape vectors $\alpha,\beta$
 & $\alpha_k,\beta_k>0$; support is $(0,1)^n$.\\[3pt]
Dirichlet & Concentration vector $\alpha$
 & $\alpha_k>0$; the joint support is the simplex.\\[3pt]
Weibull & Shape $p$ and scale $s$
 & The powers $p_k>0$ are fixed when known;
   $s_k>0$ is learned.\\[3pt]
Pareto & Lower bound $m$ and shape $\kappa$
 & $m$ is fixed by the action domain; $\kappa_k>0$, with stronger bounds
   when higher moments are required.\\[3pt]
Log-Normal & Logarithmic mean $\mu$ and covariance $\Sigma=BB^\top$
 & $B$ has a positive diagonal; the density includes its Jacobian factor.\\
\bottomrule
\end{tabular}
\end{table}

Valid distribution parameters ensure normalization and the prescribed
action support. Admissibility additionally requires the state
regularity, entropy integrability, and moment bounds in
Section~\ref{Section 2}. These conditions must be checked for the chosen
parameter functions. 

The role of the Hamiltonian coefficients can be seen directly in a
one-dimensional Gamma policy. Its parameter relations are
$\alpha=1+K/\lambda$ and $\beta=-L/\lambda$. Thus
\begin{equation}\label{eq:sec4_gamma_q}
q_\omega(t,x,i;a)
=\lambda\bigl[
\alpha_\omega\ln\beta_\omega-\ln\Gamma(\alpha_\omega)
+(\alpha_\omega-1)\ln a-\beta_\omega a\bigr].
\end{equation}
The action-dependent coefficients and the normalizing term are linked
by the distributional formula. They are not fitted independently.
If $K$ is known, the shape $1+K/\lambda$ is inserted directly into
\eqref{eq:sec4_parameter_map}. Analogous relations apply to the other
families. The temperature is kept fixed during a training run.

\begin{example}[Gamma--Gaussian policy for consumption and investment]
\label{ex:sec4_consumption_policy}
In the model of Section~\ref{Section 5}, the Hamiltonian separates into
a logarithmic-linear consumption term and a quadratic investment term.
The resulting policy can be parameterized as
\begin{equation}\label{eq:sec4_consumption_policy}
\begin{aligned}
\boldsymbol{\pi}_\omega(a_1,a_2\mid t,x,i)
={}&\mathrm{Gamma}
 \bigl(a_1\mid\alpha_c(t),\beta_{c,\omega}(t,x,i)\bigr)\\
&\times\mathcal N
 \bigl(a_2\mid\mu_{p,\omega}(t,x,i),s_{p,\omega}^2(t,x,i)\bigr),\\
\alpha_c(t)={}&1+\frac{e^{-\rho t}}{\lambda}.
\end{aligned}
\end{equation}
The network learns $\beta_{c,\omega}>0$, $\mu_{p,\omega}\in\mathbb R$,
and $s_{p,\omega}>0$. The consumption shape and the independence of the
two action components are supplied by the model structure. Evaluating
this policy requires neither the market coefficients nor derivatives of
the value function.
\end{example}

This construction gives the results of Section~\ref{Section 3} three
computational roles: they determine the approximation family, provide
normalized densities for sampling and evaluation, and fix known
parameter relations. In particular, $q_\omega$ and the policy use the
same parameters. No separate action-dependent network or numerical
action-space integration is needed to normalize these explicit families.

\subsection{Martingale characterization of the $q$-Function}\label{sub42}
We use the discretely sampled state--action construction of
\citet{jia2025erratum}. The payoff $J(\cdot;\boldsymbol\pi)$ and its
$q$-function remain defined through the exploratory problem in
Section~\ref{Section 2}. The observed trajectories, however, are generated
by the original controlled SDE with sampled actions held fixed between
sampling times.

For an initial time $t$, fix a deterministic grid
$\mathcal G=\{t=t_0<t_1<\cdots<t_K=T\}$. Enlarge the probability space
with independent uniform sampling variables $U_0,\ldots,U_{K-1}$ on
$(0,1)$, independent of $W$ and $\theta$.
Given a behavior policy $\boldsymbol\nu$, draw
\[
A_k^{\mathcal G,\boldsymbol\nu}\sim
\boldsymbol\nu(\cdot\mid t_k,X_{t_k}^{\mathcal G,\boldsymbol\nu},\theta_{t_k})
\]
using $U_k$, and hold this action until $t_{k+1}$. The resulting state
process satisfies
\begin{equation}\label{eq:sampled_state_sde}
\begin{aligned}
dX_s^{\mathcal G,\boldsymbol\nu}
={}&b(s,X_s^{\mathcal G,\boldsymbol\nu},\theta_{s-};
       A_k^{\mathcal G,\boldsymbol\nu})\,ds+\sigma(s,X_s^{\mathcal G,\boldsymbol\nu},\theta_{s-};
       A_k^{\mathcal G,\boldsymbol\nu})\,dW_s,
\quad s\in(t_k,t_{k+1}],\\
X_t^{\mathcal G,\boldsymbol\nu}={}&x,\quad\theta_t=i.
\end{aligned}
\end{equation}
Write $A_s^{\mathcal G,\boldsymbol\nu}=A_k^{\mathcal G,\boldsymbol\nu}$
on $[t_k,t_{k+1})$. Let $\mathbb F^{\mathcal G}$ be the usual
augmentation of the filtration generated by $W$, $\theta$, and the
sampling variables revealed up to the current time. The action on each
interval is measurable immediately after its sampling time.

For the policies and grids considered below, assume the sampled process
has the moments needed for the polynomial growth of $\hat J$ and that,
for some $p\geq1$ satisfying $|\hat J(s,y,j)|\leq C(1+|y|^p)$,
\begin{equation}\label{eq:sampled_integrability}
\begin{aligned}
\E_{t,x,i}\left[
\sup_{t\leq s\leq T}|X_s^{\mathcal G,\boldsymbol\nu}|^p
+\int_t^T\left(
|r(s,X_s^{\mathcal G,\boldsymbol\nu},\theta_s;
          A_s^{\mathcal G,\boldsymbol\nu})|+|\hat q(s,X_s^{\mathcal G,\boldsymbol\nu},\theta_s;
          A_s^{\mathcal G,\boldsymbol\nu})|
\right)ds\right]<\infty.
\end{aligned}
\end{equation}

\begin{lemma}[Martingale Characterization Theorem]\label{Theorem Martingale Characterization}
Under Assumption~\ref{Assumption 1} and condition~\eqref{eq:sampled_integrability}, let a policy $\boldsymbol{\pi} \in \boldsymbol{\Pi}$, a function $\hat{J}(\cdot, \cdot, i) \in C^{1,2}([0, T) \times \mathbb{R}^d) \cap C([0, T] \times \mathbb{R}^d)$ with polynomial growth for each $i \in \mathcal{S}$, and a continuous function $\hat{q}: [0, T) \times \mathbb{R}^d \times \mathcal{S} \times \mathcal{A} \rightarrow \mathbb{R}$ be given satisfying
\begin{equation*}
\left\{\begin{array}{ll}
\hat{J}(T, x, i) = g(x, i),\\
\displaystyle\int_{\mathcal{A}} \left[ \hat{q}(t, x, i; a) - \lambda \ln \boldsymbol{\pi}(a \mid t, x, i) \right] \boldsymbol{\pi}(a \mid t, x, i) \,da = 0,
\end{array} \right.
\end{equation*}
for all $(t, x, i) \in [0, T) \times \mathbb{R}^d \times \mathcal{S}$. A policy has full support if, at every $(t,x,i)$ with $t<T$, it assigns positive probability to every nonempty relatively open subset of $\mathcal A$. Then
\begin{itemize}
\item[(\romannumeral 1)] If $\boldsymbol{\pi}$ has full support, $\hat{J}$ and $\hat{q}$ are respectively the exploratory payoff function and the $q$-function associated with $\boldsymbol{\pi}$ if and only if for all $(t, x, i) \in [0, T) \times \mathbb{R}^d \times \mathcal{S}$ and every sampling grid $\mathcal G$, the following process
\begin{equation}\label{eq:martingale_pi}
    \begin{aligned}
    &\hat{J}(s, X_s^{\mathcal G,\boldsymbol{\pi}}, \theta_s)+\int_t^s\Bigl[
    r(\tau, X_\tau^{\mathcal G,\boldsymbol{\pi}}, \theta_\tau; A_\tau^{\mathcal G,\boldsymbol{\pi}})-\hat{q}(\tau, X_\tau^{\mathcal G,\boldsymbol{\pi}}, \theta_\tau; A_\tau^{\mathcal G,\boldsymbol{\pi}})
    \Bigr]\,d\tau
    \end{aligned}
    \end{equation}
    is an $(\mathbb F^{\mathcal G},\mathbb P)$-martingale, where $\{X_s^{\mathcal G,\boldsymbol{\pi}}, t \leq s \leq T\}$ is the solution to \eqref{eq:sampled_state_sde} under $\boldsymbol{\pi}$ with $X_t^{\mathcal G,\boldsymbol{\pi}} = x$ and $\theta_t = i$.
\item[(\romannumeral 2)] If $\hat{J}$ and $\hat{q}$ are respectively the exploratory payoff function and the $q$-function associated with $\boldsymbol{\pi}$, then for any $\boldsymbol{\pi}' \in \boldsymbol{\Pi}$ and all $(t, x, i) \in [0, T) \times \mathbb{R}^d \times \mathcal{S}$ and every sampling grid $\mathcal G$, the following process
\begin{equation}\label{eq:martingale_pi_prime}
    \begin{aligned}
    &\hat{J}(s, X_s^{\mathcal G,\boldsymbol{\pi}'}, \theta_s)+\int_t^s\Bigl[
    r(\tau, X_\tau^{\mathcal G,\boldsymbol{\pi}'}, \theta_\tau; A_\tau^{\mathcal G,\boldsymbol{\pi}'})-\hat{q}(\tau, X_\tau^{\mathcal G,\boldsymbol{\pi}'}, \theta_\tau; A_\tau^{\mathcal G,\boldsymbol{\pi}'})
    \Bigr]\,d\tau
    \end{aligned}
    \end{equation}
    is an $(\mathbb F^{\mathcal G},\mathbb P)$-martingale, where $\{X_s^{\mathcal G,\boldsymbol{\pi}'}, t \leq s \leq T\}$ is the solution to \eqref{eq:sampled_state_sde} under $\boldsymbol{\pi}'$ with initial condition $X_t^{\mathcal G,\boldsymbol{\pi}'} = x$ and $\theta_t = i$.
\item[(\romannumeral 3)] If there exists $\boldsymbol{\pi}' \in \boldsymbol{\Pi}$ with full support such that for all $(t, x, i) \in [0, T) \times \mathbb{R}^d \times \mathcal{S}$ and every sampling grid $\mathcal G$, \eqref{eq:martingale_pi_prime} is an $(\mathbb F^{\mathcal G},\mathbb P)$-martingale where $X_t^{\mathcal G,\boldsymbol{\pi}'} = x$ and $\theta_t = i$, then $\hat{J}$ and $\hat{q}$ are respectively the exploratory payoff function and the $q$-function associated with $\boldsymbol{\pi}$.
\end{itemize}
\end{lemma}
\begin{proof}
We omit the proof here, as it follows the same arguments as those used for Theorem 3.5 of \citet{zhang2026continuous}.
\end{proof}
\begin{remark}
Lemma \ref{Theorem Martingale Characterization}-(\romannumeral 1) facilitates \textit{on-policy} learning, where learning the $q$-function of a given target policy $\boldsymbol{\pi}$ is based on the data $\{(s, X_s^{\mathcal G,\boldsymbol{\pi}}, \theta_s, A_s^{\mathcal G,\boldsymbol{\pi}}), t \leq s \leq T\}$ generated by $\boldsymbol{\pi}$ itself. On the other hand, Lemma \ref{Theorem Martingale Characterization}-(\romannumeral 2) and -(\romannumeral 3) provide the theoretical foundation for \textit{off-policy} learning, where learning the $q$-function of $\boldsymbol{\pi}$ is based on data generated by a different policy $\boldsymbol{\pi}'$, known as the behavior policy.
\end{remark}

\subsection{Algorithm design}
For brevity, suppress its grid and behavior
policy superscripts and write it as $X$. Let
$
\mathcal R_k
=
\int_{t_k}^{t_{k+1}}
r(s,X_s,\theta_{s-};A_k)\,ds
$
be the observed accumulated reward over a sampling interval.
We use the residual
\begin{equation}\label{theoretical_TD}
\begin{aligned}
\delta_k(\phi,\omega)
={}&V_\phi(t_{k+1},X_{t_{k+1}},\theta_{t_{k+1}})
    -V_\phi(t_k,X_{t_k},\theta_{t_k})\\
&+\mathcal R_k
    -q_\omega(t_k,X_{t_k},\theta_{t_k};A_k)\Delta t.
\end{aligned}
\end{equation}
If the environment supplies only the instantaneous reward rate $r_k$,
replace $\mathcal R_k$ by $r_k\Delta t$. At the terminal step,
$V_\phi(T,X_T,\theta_T)=g(X_T,\theta_T)$.
The product
$q_\omega(t_k,X_{t_k},\theta_{t_k};A_k)\Delta t$
approximates
$\int_{t_k}^{t_{k+1}}q_\omega(s,X_s,\theta_s;A_k)\,ds$. Thus the continuous integral gives
the exact martingale relation at the true functions; the endpoint
approximation generally introduces discretization bias at a fixed
$\Delta t$.

For an exact martingale increment $dM_s$, a predictable test process
$\xi$ satisfies $\E\int\xi_s\,dM_s=0$ whenever the stochastic integral
is a true integrable martingale, for example under the usual
square-integrability condition.
We approximate these orthogonality conditions using the tests
\begin{equation}\label{eq:sec4_test_functions}
\begin{aligned}
\xi_k
&=\nabla_\phi V_\phi(t_k,X_{t_k},\theta_{t_k}),\\
\zeta_k
&=\nabla_\omega q_\omega(t_k,X_{t_k},\theta_{t_k};A_k)\\
&=\lambda\nabla_\omega
  \ln\boldsymbol{\pi}_\omega
  (A_k\mid t_k,X_{t_k},\theta_{t_k}).
\end{aligned}
\end{equation}
Each test is evaluated before the next transition is observed and is
held fixed over that interval. The sampled action is treated as fixed
when taking the parameter derivative. A stochastic approximation step is
\begin{equation}\label{eq:sec4_updates}
\phi\leftarrow\phi+\gamma_\phi\xi_k\delta_k,\qquad
\omega\leftarrow\omega+\gamma_\omega\zeta_k\delta_k,
\end{equation}
with positive learning rates. Both increments are computed at the same
pre-update parameter pair. The constant $\lambda$ in $\zeta_k$ can
equivalently be absorbed into $\gamma_\omega$.

Algorithm~\ref{alg:structure_q_learning} uses the current structured
policy to collect data. Its incremental version updates after each
transition; its episodic version freezes the parameters during an
episode and applies the accumulated increments at the end. This
distinction concerns update timing, rather than whether the data are
on-policy or off-policy. Transitions from a different behavior policy
can also be used with the same normalized $q$-residual under the support
and coverage conditions discussed above.

\begin{algorithm}[h!]
\caption{Structure-informed $q$-learning}
\label{alg:structure_q_learning}
\small
\begin{algorithmic}[1]
\State \textbf{Input:} Initial state and regime $(x_0,i_0)$,
horizon $T$, grid size $K$, episodes $N$, temperature $\lambda$,
and update mode.
\State Select $\mathcal D$ and $\mathcal T_{\mathcal D}$ from the
action domain and Hamiltonian structure; fix all known parameters.
\State Initialize $V_\phi$ with terminal value $g$ and define
$(\boldsymbol{\pi}_\omega,q_\omega)$ by
\eqref{eq:sec4_shared_parameterization}.
\State Set $\Delta t=T/K$  and positive learning-rate schedules
$\gamma_{\phi,j},\gamma_{\omega,j}$.
\For{$j=1,\ldots,N$}
    \State Reset the environment to
    $(X_0,\theta_0)=(x_0,i_0)$.
    \State Set $\Delta\phi=0$ and $\Delta\omega=0$.
    \For{$k=0,\ldots,K-1$}
        \State Evaluate
        $\eta_\omega(t_k,X_{t_k},\theta_{t_k})$.
        \State Sample
        $A_k\sim f_{\mathcal D}
        (\cdot;\eta_\omega(t_k,X_{t_k},\theta_{t_k}))$.
        \State Hold $A_k$ over $[t_k,t_{k+1})$; observe
        $X_{t_{k+1}},\theta_{t_{k+1}},\mathcal R_k$.
        \State Evaluate
        $q_\omega(t_k,X_{t_k},\theta_{t_k};A_k)$
        using the explicit log-density of $\mathcal D$.
        \State Compute $\delta_k$ from \eqref{theoretical_TD},
        using $g(X_T,\theta_T)$ at the terminal step.
        \State Compute $\xi_k$ and $\zeta_k$ from
        \eqref{eq:sec4_test_functions}
        at the pre-update parameters.
        \If{the update mode is incremental}
            \State Apply \eqref{eq:sec4_updates}
            with learning rates
            $\gamma_{\phi,j},\gamma_{\omega,j}$.
        \Else
            \State
            $\Delta\phi\leftarrow
            \Delta\phi+\xi_k\delta_k$,
            \quad
            $\Delta\omega\leftarrow
            \Delta\omega+\zeta_k\delta_k$.
        \EndIf
    \EndFor
    \If{the update mode is episodic}
        \State
        $\phi\leftarrow\phi+\gamma_{\phi,j}\Delta\phi$,
        \quad
        $\omega\leftarrow\omega+\gamma_{\omega,j}\Delta\omega$.
    \EndIf
\EndFor
\State \textbf{Output:} $V_\phi$, $\boldsymbol{\pi}_\omega$.
\end{algorithmic}
\end{algorithm}

\section{Application to consumption-investment problem}\label{Section 5}
We apply the framework to consumption and investment in a regime-switching financial market; see \citet{merton1971optimum} and \citet{heunis2015utility} for the underlying control models. The exploratory Hamiltonian yields independent Gamma consumption and Gaussian investment distributions. We first establish the derivative signs required by these densities and then use a logarithmic terminal-utility benchmark to assess the learning method in Section~\ref{Section 4}.

\subsection{Model}
We consider a consumption-investment problem in a financial market governed by regime switching. This corresponds to a special case of our framework where the state variable is one-dimensional, and both running payoff and terminal payoff are considered.

The financial market consists of a risk-free asset (e.g., a bank account) and a risky asset (e.g., a stock). The market parameters depend on a continuous-time Markov chain $\theta = \{\theta_s, t \leq s \leq T\}$ taking values in a finite state space $\mathcal{S} = \{1, 2, \ldots, m\}$, representing the market regimes (e.g., bull and bear markets). For each regime $i \in \mathcal{S}$, the risk-free interest rate is $R(i) > 0$, the expected return of the risky asset is $\mu(i) > R(i)$, and the volatility is $\sigma(i) > 0$. 

An investor continuously decides her consumption rate $c_s \in (0, \infty)$ and the proportion of wealth invested in the risky asset $u_s \in \mathbb{R}$. The joint control is $(c_s, u_s) \in \mathcal{A}$, where $\mathcal A=(0,\infty)\times\mathbb R$. Let $X_s$ denote the investor's wealth process. The classical wealth dynamics are given by the following SDE
\begin{equation}
\left\{\begin{array}{ll}
d X_s=\left[X_{s} R(\theta_{s}) + u_s X_{s} (\mu(\theta_{s}) - R(\theta_{s})) - c_s\right] d s+u_s X_{s} \sigma(\theta_{s}) d W_s, \quad s \in(t, T],\\
X_t=x > 0, \quad \theta_t=i,
\end{array} \right.
\end{equation}
where $W_s$ is a standard one-dimensional Brownian motion independent of the Markov chain $\theta_s$.

The investor aims to maximize the expected discounted utility of intermediate consumption and terminal wealth. We adopt logarithmic utility for both intermediate consumption and terminal wealth. The classical payoff function is
\begin{equation}
J^{cl}(t, x, i ; c, u)
=\E_{t,x,i}\left[\int_t^T e^{-\rho s} \ln(c_s) d s+ e^{-\rho T} \ln(X_T)\right] ,
\end{equation}
where $\rho > 0$ is the discount rate.

Now we introduce the ERRL optimal control framework. The deterministic action $(c_s, u_s)$ is replaced by a stochastic feedback policy $\boldsymbol{\pi}(c, u \mid t, x, i)$. Under the exploratory framework, the state dynamics are governed by the corresponding regime-switching exploratory SDE
\begin{equation}\label{exploratory SDE Merton}
\left\{\begin{array}{ll}
d X_s=&\displaystyle \left(\int_{\mathcal{A}} \left[X_{s} R(\theta_{s}) + u X_{s} (\mu(\theta_{s}) - R(\theta_{s})) - c\right] \boldsymbol{\pi}(c, u \mid s,X_{s},\theta_{s}) \,dc\,du\right) d s \\
&\displaystyle +\left\{\int_{\mathcal{A}} u^2 X_{s}^2 \sigma^2(\theta_{s}) \boldsymbol{\pi}(c, u \mid  s,X_{s},\theta_{s}) \,dc\,du\right\}^{1/2} d W_s, \quad s \in(t, T],\\
X_t=&x > 0, \quad \theta_t=i.
\end{array} \right.
\end{equation}
The exploratory payoff function is
\begin{equation}\label{payoff function Merton}
\begin{aligned}
J(t, x, i; {\boldsymbol{\pi}})
=&\E_{t,x,i}\bigg[{\int}_t^T \Big( \int_{\mathcal{A}} e^{-\rho s} \ln(c) \boldsymbol{\pi}(c, u \mid  s,X_{s},\theta_{s}) \,dc\,du \\
& -\lambda{\int}_{\mathcal{A}} \boldsymbol{\pi}(c, u \mid  s,X_{s},\theta_{s})\ln \boldsymbol{\pi}(c, u \mid  s,X_{s},\theta_{s}) \,dc\,du \Big)ds +e^{-\rho T} \ln(X_T)\bigg].
\end{aligned}
\end{equation}

\subsection{Exploratory optimal strategy}\label{subsec:merton-policy}
\begin{proposition}\label{proposition increasing and concave Merton}
Suppose that, for every $i\in\mathcal S$,
$V(\cdot,\cdot,i)\in C^{1,2}([0,T)\times(0,\infty))$
is finite-valued and satisfies the exploratory HJB equation
for the consumption-investment problem,
with pointwise maximization over probability densities on
$\mathcal A=(0,\infty)\times\mathbb R$. Then
\[
V_x(t,x,i)>0,
\quad
V_{xx}(t,x,i)<0,
\quad
(t,x,i)\in[0,T)\times(0,\infty)\times\mathcal S.
\]
Consequently, for each $t\in[0,T)$ and $i\in\mathcal S$,
the exploratory value function $V(t,\cdot,i)$ is strictly
increasing and strictly concave.
\end{proposition}

The proof can be found in Appendix~\ref{Appendix D}.

\begin{theorem}
The optimal exploratory strategy $\boldsymbol{\pi}^*(c, u \mid t,x,i)$ is completely separable, i.e., $\boldsymbol{\pi}^*(c, u \mid t,x,i) = \boldsymbol{\pi}_{c}^*(c \mid t,x,i) \boldsymbol{\pi}_{u}^*(u \mid t,x,i)$, where the optimal distributions are given by
\begin{equation}\label{Optimal exploratory strategy Merton}
\left\{\begin{array}{ll}
\boldsymbol{\pi}_{u}^*(u \mid t,x,i) = \mathcal{N} \left(u \mid-\dfrac{(\mu(i) - R(i))V_x}{x \sigma^2(i) V_{xx}}, -\dfrac{\lambda}{x^2 \sigma^2(i) V_{xx}}\right),\\
\boldsymbol{\pi}_{c}^*(c \mid t,x,i) =\mathrm{Gamma}\left(c \mid\frac{e^{-\rho t}}{\lambda} + 1,\frac{V_x}{\lambda}\right).
\end{array} \right.
\end{equation}
\end{theorem}

\begin{proof}
According to Theorem \ref{Theorem Optimal strategy}, the optimal exploratory strategy is 
\begin{equation*}
\boldsymbol{\pi}^*(c, u\mid t,x,i)=\frac{\displaystyle\exp \left(\frac{1}{\lambda}H(t,x,i,V_x,V_{xx};c, u)\right)}{\displaystyle\int_\mathcal{A} \exp \left(\frac{1}{\lambda}H(t,x,i,V_x,V_{xx};c, u)\right) \,dc\,du}.
\end{equation*}
In this model, the Hamiltonian is given by
\begin{equation*}
\begin{aligned}
&H(t, x, i, V_x, V_{xx}; c, u)\\ =& e^{-\rho t} \ln(c) + \left[x R(i) + u x (\mu(i) - R(i)) - c\right] V_x + \frac{1}{2} u^2 x^2 \sigma^2(i) V_{xx} \\
=& \left( e^{-\rho t} \ln(c) - c V_x \right) + \left( \frac{1}{2} x^2 \sigma^2(i) V_{xx} u^2 + x (\mu(i) - R(i)) V_x u \right) + x R(i) V_x.
\end{aligned}
\end{equation*}
Since the terms involving the consumption rate $c$ and the portfolio proportion $u$ are entirely decoupled in the Hamiltonian, the measure $\exp(\frac{1}{\lambda}H)$ naturally factors into two independent distributions.

For the portfolio selection $u \in \mathbb{R}$, the corresponding density function is proportional to
\begin{equation*}
\exp\left( \frac{1}{\lambda} \left[ \frac{1}{2} x^2 \sigma^2(i) V_{xx} u^2 + x (\mu(i) - R(i)) V_x u \right] \right).
\end{equation*}
By Proposition \ref{proposition increasing and concave Merton}, the value function is strictly concave with respect to the wealth variable, which guarantees $V_{xx} < 0$. This strict concavity is crucial, as it ensures that the coefficient of the quadratic term $u^2$ in the exponent is strictly negative. Consequently, the exponential function represents a downward-opening parabola in the exponent, guaranteeing that the density is integrable over $\mathbb{R}$. Completing the square for $u$ yields the Gaussian distribution $\mathcal{N}(\mu_{u}, \sigma^2_{u})$ with mean $\mu_{u} = -\frac{(\mu(i) - R(i))V_x}{x \sigma^2(i) V_{xx}}$ and variance $\sigma^2_{u} = -\frac{\lambda}{x^2 \sigma^2(i) V_{xx}} > 0$.

Simultaneously, for the consumption rate $c \in [0, \infty)$, the density is proportional to
\begin{equation*}
\exp\left( \frac{1}{\lambda} \left[ e^{-\rho t} \ln(c) - c V_x \right] \right) = c^{\frac{e^{-\rho t}}{\lambda}} \exp\left(-\frac{V_x}{\lambda}c\right).
\end{equation*}
According to Proposition \ref{proposition increasing and concave Merton}, the value function is strictly increasing, which ensures that the marginal value of wealth is strictly positive, i.e., $V_x > 0$. This positivity property is essential for the legitimacy of the distribution, as it guarantees that the exponential decay term $\exp(-\frac{V_x}{\lambda}c)$ strongly dominates the polynomial term $c^{\frac{e^{-\rho t}}{\lambda}}$ as $c \to \infty$. Therefore, the integral is perfectly convergent on $[0, \infty)$. We can clearly recognize this algebraic structure as the kernel of a standard Gamma distribution $\mathrm{Gamma}(\alpha, \beta)$ up to a normalizing constant, where the shape parameter is $\alpha = \frac{e^{-\rho t}}{\lambda} + 1 > 0$ and the rate parameter is $\beta = \frac{V_x}{\lambda} > 0$.
\end{proof}
\subsection{Numerical experiments}\label{subsec:numerics}
We examine three questions: whether the learned policy and value function agree with an independently computable benchmark, how much fixing a known distributional parameter changes learning speed, and how the time grid affects the reference solution. All reported learning results are obtained from simulated transition data.\footnote{The code for the numerical experiments is available at \url{https://github.com/hcdmath/A-Study-on-Distributional-Classes.git}.} The benchmark coefficients are used by the simulator and for evaluation, but are not passed to the learner.

\subsubsection{Logarithmic benchmark}
Throughout this subsection, $\mathrm{Gamma}(\alpha,\beta)$ denotes a
Gamma distribution with shape $\alpha$ and rate $\beta$, and the second
parameter of $\mathcal N$ denotes the variance. Define
$\alpha(t)=1+e^{-\rho t}/\lambda$.
For the logarithmic utility specified above, the value function takes the form
\begin{equation}\label{eq:log-benchmark-value}
V(t,x,i)=A(t)\ln x+B_i(t),\quad
A(t)=e^{-\rho T}+\frac{e^{-\rho t}-e^{-\rho T}}{\rho}+\lambda(T-t).
\end{equation}
Indeed, the coefficient of $\ln x$ gives $A'(t)+e^{-\rho t}+\lambda=0$, with $A(T)=e^{-\rho T}$. The remaining coefficients satisfy the linear system
\begin{equation}\label{eq:log-benchmark-B}
B'(t)+QB(t)+F(t)=0,\quad B(T)=0,
\end{equation}
where $Q$ is the generator of the regime process,
$B(t)=(B_1(t),\ldots,B_m(t))^\top$, and
$F(t)=(F_1(t),\ldots,F_m(t))^\top$, with
\begin{equation}\label{eq:log-benchmark-F}
\begin{aligned}
F_i(t)={}&A(t)\left(R(i)+\frac{(\mu(i)-R(i))^2}{2\sigma^2(i)}\right)
 +\lambda\ln\Gamma(\alpha(t))
 +(e^{-\rho t}+\lambda)\ln\!\left(\frac{\lambda}{A(t)}\right)\\
&+\frac{\lambda}{2}\ln\!\left(\frac{2\pi\lambda}{A(t)\sigma^2(i)}\right).
\end{aligned}
\end{equation}
Here $\Gamma$ denotes the Gamma function. Thus $B(t)=\int_t^T e^{Q(s-t)}F(s)\,ds$. The associated policy is
\begin{equation}\label{eq:log-benchmark-policy}
c\sim\mathrm{Gamma}\!\left(1+\frac{e^{-\rho t}}{\lambda},\frac{A(t)}{\lambda x}\right),
\quad
u\sim\mathcal N\!\left(\frac{\mu(i)-R(i)}{\sigma^2(i)},\frac{\lambda}{A(t)\sigma^2(i)}\right),
\end{equation}
with independent components. Its consumption mean divided by wealth equals $(e^{-\rho t}+\lambda)/A(t)$. Since $A(t)\ge e^{-\rho T}>0$, the averaged wealth equation under this policy is geometric, with bounded time-dependent coefficients in each regime. Wealth remains positive and the required logarithmic moments are finite. Direct evaluation of its payoff gives \eqref{eq:log-benchmark-value}. In this particular logarithmic model the regime-transition rates affect $B$ but not the maximizing policy.

We use $T=1$, $\rho=0.05$, $\lambda=0.10$, and the two-regime parameters in Table~\ref{tab:market}. The system \eqref{eq:log-benchmark-B} is integrated backward using DOP853 with relative tolerance $2\times10^{-12}$ and absolute tolerance $2\times10^{-13}$. It gives
\[
A(0)=2.0266409345,\qquad
B_1(0)=-1.2451750848,\qquad B_2(0)=-1.3223975466.
\]
An independent check substitutes finite-difference approximations of the
benchmark value derivatives into the exploratory HJB equation after
maximization over the policy densities. At twelve interior time--wealth
points in each regime, the maximum absolute residual is
$1.55\times10^{-8}$. Quadrature checks of the policy densities have normalization error below $3\times10^{-15}$.

\begin{table}[h!]
\centering
\caption{Market parameters for the numerical benchmark. Rates use the same time unit as $T$.}
\label{tab:market}
\begin{tabular}{ccccc}
\toprule
Regime $i$ & $R(i)$ & $\mu(i)$ & $\sigma(i)$ & $(\mu(i)-R(i))/\sigma^2(i)$\\
\midrule
1 & 0.02 & 0.08 & 0.20 & 1.500000\\
2 & 0.01 & 0.04 & 0.35 & 0.244898\\
\bottomrule
\end{tabular}
\qquad
$Q=\begin{pmatrix}-0.8&0.8\\0.5&-0.5\end{pmatrix}$.
\end{table}

\subsubsection{Parameterization and sampling protocol}
We implement a batched off-policy version of the residual and test-function updates in Section~\ref{Section 4}. Write $y=\ln x$ and let
$f(t)=(P_0(2t/T-1),\ldots,P_3(2t/T-1))^\top$,
where $P_k$ denotes the Legendre polynomial of degree $k$. Both learners use the same twelve-parameter critic
\begin{equation}\label{eq:numerical-critic}
V_\phi(t,e^y,i)=e^{-\rho T}y
 +(T-t)\bigl[(a^\top f(t))y+b_i^\top f(t)\bigr],
\qquad a,b_1,b_2\in\mathbb R^4.
\end{equation}
The critic parameter vector is $\phi=(a^\top,b_1^\top,b_2^\top)^\top$. This parameterization enforces the terminal condition exactly. The logarithmic wealth dependence is shared by both methods.

For the actor, write $C=c/x$. Let $\eta(t)=\eta^\top f(t)$ and
$\ell_i(t)=v_i^\top f(t)$, where $\eta,v_1,v_2\in\mathbb R^4$
are learned coefficient vectors, and let $m_i\in\mathbb R$ be a learned
portfolio mean for each regime. The symbol $\omega$ collects all actor
parameters. The policy has independent components
\begin{equation}\label{eq:numerical-actor}
C\sim\mathrm{Gamma}(\alpha_\omega(t),\beta_\omega(t)),
\quad \beta_\omega(t)=\alpha_\omega(t)e^{\eta(t)},
\qquad u\sim\mathcal N(m_i,e^{\ell_i(t)}).
\end{equation}
The absolute-consumption density includes the change of variables:
\begin{equation}\label{eq:numerical-q}
q_\omega(t,x,i;c,u)
=\lambda\bigl[\ln f_\Gamma(c/x;\alpha_\omega,\beta_\omega)
              +\ln f_N(u;m_i,e^{\ell_i})-\ln x\bigr].
\end{equation}
Here $f_\Gamma(\cdot;\alpha,\beta)$ and $f_N(\cdot;m,s^2)$ denote
the Gamma and Gaussian densities, respectively, and time arguments are
suppressed in the density parameters. The joint policy density is
$\boldsymbol\pi_\omega(c,u\mid t,x,i)=\exp(q_\omega(t,x,i;c,u)/\lambda)$.
Consequently, the normalization constraint holds exactly. The fixed-shape method sets $\alpha_\omega(t)=\alpha(t)=1+e^{-\rho t}/\lambda$ and has fourteen actor parameters: four for $\eta$, two portfolio means, and eight for the log variances. The learned-shape ablation sets
$\alpha_\omega(t)=\alpha(t)\exp(z^\top f(t))$, with four additional parameters $z\in\mathbb R^4$. Both methods use the same approximation spaces for the unknown consumption mean and portfolio moments; the fixed-shape class is the $z=0$ subfamily of the learned-shape class. Their total parameter counts are 26 and 30. This comparison tests the benefit of fixing a known parameter.

Both actors start from $\E[C]=0.8$, $m_1=m_2=0.75$, and portfolio variance $0.8$. The learned-shape method starts at $z=0$, so the initial policies are identical. The critic starts at $a=(0.5,0,0,0)^\top$ and $b_1=b_2=0$.

For each batch, the simulator independently samples $t$ uniformly from $\{0,h,\ldots,T-h\}$, $y$ uniformly from $[-1,1]$, and the regime uniformly from $\{1,2\}$. The behavior actions are $C\sim\mathrm{Gamma}(4,4)$ and $u\sim\mathcal N(0,4)$, independently. This behavior density has full support on the action domain. A next regime $j$ is sampled from row $i$ of $e^{Qh}$. Writing $Y=\ln X$, for each sampled state and action we generate two transitions using opposite normal shocks:
\begin{equation}\label{eq:numerical-transition}
Y^\pm_{t+h}=y+
 \left(R(i)+(\mu(i)-R(i)) u-C-\tfrac12\sigma^2(i) u^2\right)h
 \ \pm\sigma(i) u\sqrt h\,Z,\qquad Z\sim\mathcal N(0,1).
\end{equation}
The next regime is shared by the pair, and the observed left-endpoint reward is
$r_th=e^{-\rho t}(y+\ln C)h$. The log-wealth update preserves positivity of $X=e^Y$. It is a first-order splitting approximation to the local state generator: it freezes relative consumption and the current-regime coefficients within the step and places the regime transition at its endpoint. 

For each antithetic pair, let
\begin{equation}\label{eq:numerical-residual}
\bar\delta=\frac12\bigl[V_\phi(t+h,e^{Y^+_{t+h}},j)
                         +V_\phi(t+h,e^{Y^-_{t+h}},j)\bigr]
 -V_\phi(t,e^y,i)+r_th-q_\omega(t,e^y,i;c,u)h.
\end{equation}
The two update directions are the batch averages of
$\nabla_\phi V_\phi\,\bar\delta/h$ and
$\nabla_\omega q_\omega\,\bar\delta/h$, evaluated at the same pre-update parameters. We apply Adam moment scaling \citep{kingma2015adam}, with coefficients $(0.9,0.999)$, denominator offset $10^{-8}$, and clipping of each batch direction to Euclidean norm 10. The critic and actor step sizes at update $n$ are $0.004(1+n/1000)^{-0.55}$ and $0.003(1+n/1000)^{-0.55}$, respectively. These are stochastic-approximation updates with fixed test functions within a transition. Each update uses 1,024 pairs, or 2,048 observed transitions. Antithetic pairing cancels the direct Brownian contribution for the critic \eqref{eq:numerical-critic}; it is a simulator-based variance reduction available equally to both methods.

We use $h=0.01$, 5,000 updates per run, and ten paired seeds, the consecutive integers from 20261125 to 20261134. Each pair of methods receives identical sampled transitions and has the same initial policy. Thus each run uses 10,240,000 observed transitions, including both members of every antithetic pair. 

\subsubsection{Accuracy and the effect of fixing the shape}
Policy accuracy is measured by
\begin{equation}\label{eq:numerical-metrics}
\begin{aligned}
E_{\mathrm{KL}}&=\frac1{202}\sum_{\ell=0}^{100}\sum_{i=1}^2
D_{\mathrm{KL}}\!\left(\boldsymbol\pi^*(\cdot\mid t_\ell,1,i)
\,\Vert\,\boldsymbol\pi_\omega(\cdot\mid t_\ell,1,i)\right),\\
E_V&=\left(\frac{\sum_{(t,y,i)\in\mathcal G}
 |V_\phi(t,e^y,i)-V(t,e^y,i)|^2}
 {\sum_{(t,y,i)\in\mathcal G}|V(t,e^y,i)|^2}\right)^{1/2},
\end{aligned}
\end{equation}
where $D_{\mathrm{KL}}$ denotes the Kullback--Leibler divergence, $t_\ell=\ell T/100$, and $\mathcal G$ is the product of these 101 times, eleven equally spaced values of $y$ in $[-1,1]$, and the two regimes. Policy formulas at $T$ use their left limits. The KL divergence is evaluated analytically as the sum of Gamma and Gaussian divergences. It is independent of wealth for the homogeneous policies used here. Metrics are recorded every 50 updates. We report the first checkpoint at which $E_{\mathrm{KL}}<10^{-2}$ for three consecutive checkpoints, denoted by $N_{.01}$.

To evaluate achieved performance separately from critic approximation, we also compute the continuous-time payoff of each final actor. If its relative-consumption shape and rate are $\alpha_\omega,\beta_\omega$, and its portfolio mean and variance are $m_i$ and $s_i^2(t)=e^{\ell_i(t)}$, define
\begin{align*}
\kappa_i(t)&=R(i)+(\mu(i)-R(i)) m_i-\alpha_\omega/\beta_\omega
 -\tfrac12\sigma^2(i)(m_i^2+s_i^2),\\
h_c(t)&=\alpha_\omega-\ln\beta_\omega+\ln\Gamma(\alpha_\omega)
 +(1-\alpha_\omega)\psi(\alpha_\omega),\\
G_i^\omega(t)&=A(t)\kappa_i(t)
 +e^{-\rho t}\bigl[\psi(\alpha_\omega)-\ln\beta_\omega\bigr]
 +\lambda\bigl[h_c(t)+\tfrac12\ln(2\pi e s_i^2)\bigr],
\end{align*}
where $\psi=\Gamma'/\Gamma$ is the digamma function, $h_c(t)$ is the
differential entropy of relative consumption $C$, and the time arguments
of $\alpha_\omega$, $\beta_\omega$, and $s_i^2$ are suppressed. The policy payoff is
$J^{\pi_\omega}(t,x,i)=A(t)\ln x+B_i^\omega(t)$, with
$(B^\omega)' +QB^\omega+G^\omega=0$ and $B^\omega(T)=0$.
We report the average initial utility loss
$\Delta J=\tfrac12\sum_{i=1}^2[B_i(0)-B_i^\omega(0)]$.
This ODE uses market coefficients only for evaluation after learning.

\begin{table}[tbp]
\centering
\caption{Learning results over ten paired seeds. Entries are mean $\pm$ sample standard deviation. Accuracy and payoff loss are measured after 5,000 updates.}
\label{tab:learning-results}
\small
\begin{tabular}{lcc}
\toprule
Quantity & Fixed known shape & Learned shape\\
\midrule
Actor parameters & 14 & 18\\
$N_{.01}$ & $1320\pm25.8$ & $1510\pm21.1$\\
$E_{\mathrm{KL}}$ & $(4.76\pm2.43)\times10^{-4}$ & $(5.40\pm2.87)\times10^{-4}$\\
$E_V$ & $(4.12\pm1.17)\times10^{-3}$ & $(4.18\pm1.13)\times10^{-3}$\\
Consumption-mean relative RMSE & $(8.67\pm2.43)\times10^{-3}$ & $(9.09\pm2.95)\times10^{-3}$\\
Portfolio-mean RMSE & $(5.54\pm2.81)\times10^{-3}$ & $(5.63\pm3.03)\times10^{-3}$\\
Portfolio-variance relative RMSE & $(1.01\pm0.28)\times10^{-2}$ & $(1.00\pm0.28)\times10^{-2}$\\
$\Delta J$ & $(4.76\pm2.46)\times10^{-5}$ & $(5.40\pm2.92)\times10^{-5}$\\
Elapsed time (seconds) & $4.00\pm0.11$ & $4.21\pm0.07$\\
\bottomrule
\end{tabular}
\end{table}

Table~\ref{tab:learning-results} shows that the fixed-shape learner reaches the chosen KL threshold in 1,320 updates on average, compared with 1,510 for the learned-shape ablation, a reduction of 12.6\%. The corresponding mean transition counts are 2.70 million and 3.09 million. At 5,000 updates both methods have value relative RMSE near 0.4\%, and their final error distributions substantially overlap. The mean achieved payoffs at $(t,x)=(0,1)$ are $(-1.2452225,-1.3224453)$ for the fixed-shape method and $(-1.2452290,-1.3224516)$ for the learned-shape method. The small positive payoff losses confirm that the final actors perform close to the benchmark, independently of the fitted critic.

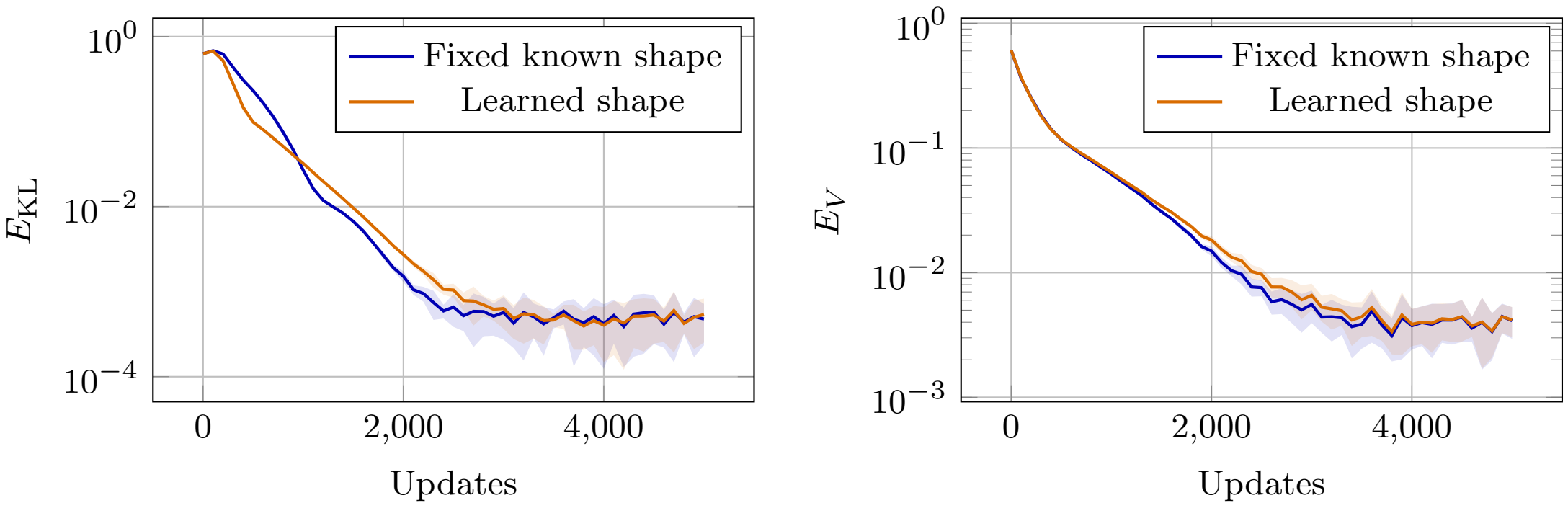
\begin{figure}[tbp]
\centering
\begin{tikzpicture}
\begin{groupplot}[group style={group size=2 by 1,horizontal sep=1.9cm},width=0.43\textwidth,height=5.1cm,xlabel={Updates},ymode=log,grid=major,legend style={font=\scriptsize,at={(0.98,0.98)},anchor=north east},tick label style={font=\scriptsize},label style={font=\small}]
\nextgroupplot[ylabel={$E_{\mathrm{KL}}$}]
\addplot[name path=fixedupper,draw=none,forget plot] coordinates {(0,0.62903128) (100,0.69126117) (200,0.64944845) (300,0.44537473) (400,0.31570639) (500,0.23733513) (600,0.16998906) (700,0.1200431) (800,0.078880955) (900,0.049215744) (1000,0.028228832) (1100,0.017312051) (1200,0.012207988) (1300,0.010357138) (1400,0.0087464785) (1500,0.0070255218) (1600,0.0055486064) (1700,0.0038783021) (1800,0.0029363184) (1900,0.0020697353) (2000,0.0017424771) (2100,0.0011888723) (2200,0.0011406984) (2300,0.0010256276) (2400,0.0007030307) (2500,0.00093351228) (2600,0.00069936574) (2700,0.00095208309) (2800,0.00086861363) (2900,0.00072451658) (3000,0.00086676235) (3100,0.00063754906) (3200,0.00098373327) (3300,0.00072615503) (3400,0.0006597347) (3500,0.00061843248) (3600,0.0007676253) (3700,0.00081719279) (3800,0.00063943628) (3900,0.00083957161) (4000,0.00071226722) (4100,0.00080548839) (4200,0.00064907655) (4300,0.00091865348) (4400,0.0009430715) (4500,0.00090883883) (4600,0.00061116802) (4700,0.00099330206) (4800,0.00056098487) (4900,0.00084842671) (5000,0.00071906889)};
\addplot[name path=fixedlower,draw=none,forget plot] coordinates {(0,0.62903128) (100,0.6719091) (200,0.59749882) (300,0.42425093) (400,0.30040886) (500,0.22531262) (600,0.16103637) (700,0.10884673) (800,0.070369173) (900,0.043883507) (1000,0.025287994) (1100,0.015352148) (1200,0.011404017) (1300,0.0094533425) (1400,0.0079545229) (1500,0.0063581168) (1600,0.0047606964) (1700,0.0035749735) (1800,0.0024079839) (1900,0.0017413311) (2000,0.0012901817) (2100,0.00091881048) (2200,0.00076072562) (2300,0.00046460887) (2400,0.00048341562) (2500,0.00038506426) (2600,0.00034303494) (2700,0.00022224155) (2800,0.00029924893) (2900,0.00030367087) (3000,0.00027215027) (3100,0.00021648182) (3200,0.00015509274) (3300,0.00028619786) (3400,0.00017613393) (3500,0.00037024787) (3600,0.00041185384) (3700,0.00013135683) (3800,0.00022314102) (3900,0.00017726206) (4000,0.00012358144) (4100,0.00024518396) (4200,0.00013022362) (4300,0.00017189219) (4400,0.00018655606) (4500,0.00024183689) (4600,0.00021924771) (4700,0.00014788545) (4800,0.00031495073) (4900,0.00016918922) (5000,0.0002333224)};
\addplot[blue!70!black,fill opacity=0.12,forget plot] fill between[of=fixedupper and fixedlower];
\addplot[blue!70!black,thick] coordinates {(0,0.62903128) (100,0.68158514) (200,0.62347363) (300,0.43481283) (400,0.30805763) (500,0.23132388) (600,0.16551271) (700,0.11444491) (800,0.074625064) (900,0.046549626) (1000,0.026758413) (1100,0.0163321) (1200,0.011806002) (1300,0.0099052404) (1400,0.0083505007) (1500,0.0066918193) (1600,0.0051546514) (1700,0.0037266378) (1800,0.0026721512) (1900,0.0019055332) (2000,0.0015163294) (2100,0.0010538414) (2200,0.00095071202) (2300,0.00074511822) (2400,0.00059322316) (2500,0.00065928827) (2600,0.00052120034) (2700,0.00058716232) (2800,0.00058393128) (2900,0.00051409372) (3000,0.00056945631) (3100,0.00042701544) (3200,0.00056941301) (3300,0.00050617645) (3400,0.00041793431) (3500,0.00049434018) (3600,0.00058973957) (3700,0.00047427481) (3800,0.00043128865) (3900,0.00050841683) (4000,0.00041792433) (4100,0.00052533618) (4200,0.00038965008) (4300,0.00054527283) (4400,0.00056481378) (4500,0.00057533786) (4600,0.00041520787) (4700,0.00057059376) (4800,0.0004379678) (4900,0.00050880796) (5000,0.00047619565)};
\addlegendentry{Fixed known shape}
\addplot[name path=freeupper,draw=none,forget plot] coordinates {(0,0.62903128) (100,0.68262411) (200,0.53488056) (300,0.30270726) (400,0.16287028) (500,0.10085995) (600,0.081312151) (700,0.066210426) (800,0.052389454) (900,0.04143541) (1000,0.032831555) (1100,0.025923809) (1200,0.020254172) (1300,0.016261247) (1400,0.012686176) (1500,0.0099638256) (1600,0.0078021101) (1700,0.0060201393) (1800,0.0047817079) (1900,0.0036122624) (2000,0.0028785627) (2100,0.0023197482) (2200,0.0019089971) (2300,0.0016556085) (2400,0.0012093683) (2500,0.0012415531) (2600,0.00097679425) (2700,0.0011494167) (2800,0.00088929965) (2900,0.00078554588) (3000,0.00090369284) (3100,0.00069261958) (3200,0.00085154045) (3300,0.00079784937) (3400,0.00067533021) (3500,0.00054657036) (3600,0.00070103865) (3700,0.00070216412) (3800,0.00058217072) (3900,0.00067726964) (4000,0.0006597141) (4100,0.00077700635) (4200,0.00073690067) (4300,0.00081292912) (4400,0.00083492261) (4500,0.00081841564) (4600,0.00064544834) (4700,0.0010092017) (4800,0.00052569674) (4900,0.00078805754) (5000,0.00082684605)};
\addplot[name path=freelower,draw=none,forget plot] coordinates {(0,0.62903128) (100,0.66759209) (200,0.50395683) (300,0.25289972) (400,0.13120944) (500,0.095871916) (600,0.079262603) (700,0.061970585) (800,0.049795801) (900,0.03918609) (1000,0.031514979) (1100,0.024388751) (1200,0.019226002) (1300,0.015091675) (1400,0.011887916) (1500,0.0092921877) (1600,0.0073414682) (1700,0.0056008356) (1800,0.004256202) (1900,0.0032794959) (2000,0.0025890637) (2100,0.0019424707) (2200,0.0015622365) (2300,0.0011090907) (2400,0.00092152824) (2500,0.00084896078) (2600,0.00059225048) (2700,0.00039987614) (2800,0.00050980251) (2900,0.00045634019) (3000,0.00036530802) (3100,0.00027820758) (3200,0.00024474611) (3300,0.0002796854) (3400,0.00023893983) (3500,0.00038296978) (3600,0.00036300844) (3700,0.0002078577) (3800,0.00020684521) (3900,0.00023445451) (4000,0.00014753225) (4100,0.00017932257) (4200,0.00012087187) (4300,0.00021685372) (4400,0.00019729956) (4500,0.00024552328) (4600,0.00025601492) (4700,0.00019217651) (4800,0.00032068989) (4900,0.00021054259) (5000,0.00025241042)};
\addplot[orange!85!black,fill opacity=0.12,forget plot] fill between[of=freeupper and freelower];
\addplot[orange!85!black,thick] coordinates {(0,0.62903128) (100,0.6751081) (200,0.51941869) (300,0.27780349) (400,0.14703986) (500,0.098365933) (600,0.080287377) (700,0.064090506) (800,0.051092628) (900,0.04031075) (1000,0.032173267) (1100,0.02515628) (1200,0.019740087) (1300,0.015676461) (1400,0.012287046) (1500,0.0096280067) (1600,0.0075717892) (1700,0.0058104875) (1800,0.004518955) (1900,0.0034458792) (2000,0.0027338132) (2100,0.0021311095) (2200,0.0017356168) (2300,0.0013823496) (2400,0.0010654483) (2500,0.0010452569) (2600,0.00078452237) (2700,0.00077464642) (2800,0.00069955108) (2900,0.00062094303) (3000,0.00063450043) (3100,0.00048541358) (3200,0.00054814328) (3300,0.00053876739) (3400,0.00045713502) (3500,0.00046477007) (3600,0.00053202355) (3700,0.00045501091) (3800,0.00039450796) (3900,0.00045586207) (4000,0.00040362317) (4100,0.00047816446) (4200,0.00042888627) (4300,0.00051489142) (4400,0.00051611108) (4500,0.00053196946) (4600,0.00045073163) (4700,0.00060068909) (4800,0.00042319332) (4900,0.00049930006) (5000,0.00053962823)};
\addlegendentry{Learned shape}
\nextgroupplot[ylabel={$E_V$}]
\addplot[name path=fixedupper,draw=none,forget plot] coordinates {(0,0.60861946) (100,0.36211705) (200,0.25411116) (300,0.18359986) (400,0.14182853) (500,0.11745254) (600,0.10194436) (700,0.089737956) (800,0.079698927) (900,0.070504368) (1000,0.062409935) (1100,0.054800286) (1200,0.048057479) (1300,0.042653918) (1400,0.036677236) (1500,0.031879129) (1600,0.027986456) (1700,0.024276989) (1800,0.020742684) (1900,0.017020044) (2000,0.016257755) (2100,0.013240235) (2200,0.011375968) (2300,0.011390834) (2400,0.0089407449) (2500,0.0087052254) (2600,0.0070596206) (2700,0.007504956) (2800,0.0072172227) (2900,0.0067609972) (3000,0.00718726) (3100,0.0055106335) (3200,0.0060445877) (3300,0.0055331132) (3400,0.0053262875) (3500,0.005260552) (3600,0.0070669502) (3700,0.0051433912) (3800,0.0043058124) (3900,0.0066779221) (4000,0.005096375) (4100,0.0053692948) (4200,0.0056358705) (4300,0.005598931) (4400,0.00567838) (4500,0.0060252394) (4600,0.0044123967) (4700,0.0063223436) (4800,0.0047376671) (4900,0.0056530611) (5000,0.0052870404)};
\addplot[name path=fixedlower,draw=none,forget plot] coordinates {(0,0.60861946) (100,0.35917689) (200,0.25161962) (300,0.18126558) (400,0.1404676) (500,0.1156346) (600,0.10015704) (700,0.08769109) (800,0.078221347) (900,0.068778766) (1000,0.06080302) (1100,0.053030126) (1200,0.046795966) (1300,0.040627697) (1400,0.034443188) (1500,0.029865984) (1600,0.026159511) (1700,0.021844681) (1800,0.018742036) (1900,0.015388765) (2000,0.013529031) (2100,0.01092692) (2200,0.0093574679) (2300,0.0080177727) (2400,0.0064244951) (2500,0.0064637356) (2600,0.0046045755) (2700,0.0046509329) (2800,0.0039374271) (2900,0.0032969818) (3000,0.0039300368) (3100,0.003287627) (3200,0.0028001854) (3300,0.0031612725) (3400,0.0020522667) (3500,0.0024537697) (3600,0.0027429097) (3700,0.0024686381) (3800,0.001940316) (3900,0.0020165083) (4000,0.0024148012) (4100,0.0025791713) (4200,0.0020567467) (4300,0.0027335528) (4400,0.0026491981) (4500,0.0027782466) (4600,0.0027753333) (4700,0.0016651528) (4800,0.0019850579) (4900,0.0032537498) (5000,0.0029446459)};
\addplot[blue!70!black,fill opacity=0.12,forget plot] fill between[of=fixedupper and fixedlower];
\addplot[blue!70!black,thick] coordinates {(0,0.60861946) (100,0.36064697) (200,0.25286539) (300,0.18243272) (400,0.14114807) (500,0.11654357) (600,0.1010507) (700,0.088714523) (800,0.078960137) (900,0.069641567) (1000,0.061606477) (1100,0.053915206) (1200,0.047426723) (1300,0.041640807) (1400,0.035560212) (1500,0.030872557) (1600,0.027072984) (1700,0.023060835) (1800,0.01974236) (1900,0.016204405) (2000,0.014893393) (2100,0.012083578) (2200,0.010366718) (2300,0.0097043031) (2400,0.00768262) (2500,0.0075844805) (2600,0.0058320981) (2700,0.0060779444) (2800,0.0055773249) (2900,0.0050289895) (3000,0.0055586484) (3100,0.0043991302) (3200,0.0044223865) (3300,0.0043471928) (3400,0.0036892771) (3500,0.0038571609) (3600,0.0049049299) (3700,0.0038060146) (3800,0.0031230642) (3900,0.0043472152) (4000,0.0037555881) (4100,0.0039742331) (4200,0.0038463086) (4300,0.0041662419) (4400,0.004163789) (4500,0.004401743) (4600,0.003593865) (4700,0.0039937482) (4800,0.0033613625) (4900,0.0044534055) (5000,0.0041158431)};
\addlegendentry{Fixed known shape}
\addplot[name path=freeupper,draw=none,forget plot] coordinates {(0,0.60861946) (100,0.36726642) (200,0.25056516) (300,0.17995988) (400,0.14065996) (500,0.11811374) (600,0.10332572) (700,0.091462857) (800,0.081645802) (900,0.07248579) (1000,0.064428391) (1100,0.05701546) (1200,0.050548462) (1300,0.045357362) (1400,0.039599223) (1500,0.035110133) (1600,0.031436799) (1700,0.027813023) (1800,0.024357518) (1900,0.020626705) (2000,0.019678731) (2100,0.016493572) (2200,0.01441082) (2300,0.014106964) (2400,0.011531778) (2500,0.010865813) (2600,0.0090307463) (2700,0.0090742043) (2800,0.0087215811) (2900,0.0079056623) (3000,0.0081538106) (3100,0.0065142107) (3200,0.0067518808) (3300,0.0061346821) (3400,0.0057600064) (3500,0.0058065419) (3600,0.0073860145) (3700,0.0054330265) (3800,0.0045125639) (3900,0.006974216) (4000,0.0051322725) (4100,0.0053472975) (4200,0.0055982943) (4300,0.0056659079) (4400,0.0056250688) (4500,0.0060553159) (4600,0.0044231844) (4700,0.0063283568) (4800,0.0047214468) (4900,0.0055769024) (5000,0.0053098433)};
\addplot[name path=freelower,draw=none,forget plot] coordinates {(0,0.60861946) (100,0.3652399) (200,0.24896327) (300,0.1783841) (400,0.13967191) (500,0.11620756) (600,0.10161289) (700,0.089647816) (800,0.080238795) (900,0.070916218) (1000,0.062978813) (1100,0.055335126) (1200,0.049240758) (1300,0.043393218) (1400,0.03748048) (1500,0.033228403) (1600,0.029648522) (1700,0.025439159) (1800,0.022362595) (1900,0.018854536) (2000,0.016922823) (2100,0.014207557) (2200,0.012212821) (2300,0.010785502) (2400,0.0088789404) (2500,0.0085508524) (2600,0.0063211361) (2700,0.0062576221) (2800,0.0052841974) (2900,0.0042245007) (3000,0.004999102) (3100,0.0040434088) (3200,0.0034938193) (3300,0.0037668711) (3400,0.0025776816) (3500,0.0030401328) (3600,0.003110797) (3700,0.002832913) (3800,0.0022098662) (3900,0.0021853995) (4000,0.0025774521) (4100,0.002666091) (4200,0.002272533) (4300,0.0028615746) (4400,0.002778265) (4500,0.002801905) (4600,0.0030519742) (4700,0.0016995353) (4800,0.0020852749) (4900,0.0032792547) (5000,0.0030435576)};
\addplot[orange!85!black,fill opacity=0.12,forget plot] fill between[of=freeupper and freelower];
\addplot[orange!85!black,thick] coordinates {(0,0.60861946) (100,0.36625316) (200,0.24976421) (300,0.17917199) (400,0.14016593) (500,0.11716065) (600,0.10246931) (700,0.090555337) (800,0.080942298) (900,0.071701004) (1000,0.063703602) (1100,0.056175293) (1200,0.04989461) (1300,0.04437529) (1400,0.038539852) (1500,0.034169268) (1600,0.03054266) (1700,0.026626091) (1800,0.023360057) (1900,0.01974062) (2000,0.018300777) (2100,0.015350565) (2200,0.01331182) (2300,0.012446233) (2400,0.010205359) (2500,0.0097083328) (2600,0.0076759412) (2700,0.0076659132) (2800,0.0070028893) (2900,0.0060650815) (3000,0.0065764563) (3100,0.0052788098) (3200,0.00512285) (3300,0.0049507766) (3400,0.004168844) (3500,0.0044233373) (3600,0.0052484058) (3700,0.0041329697) (3800,0.0033612151) (3900,0.0045798077) (4000,0.0038548623) (4100,0.0040066943) (4200,0.0039354136) (4300,0.0042637413) (4400,0.0042016669) (4500,0.0044286104) (4600,0.0037375793) (4700,0.004013946) (4800,0.0034033609) (4900,0.0044280785) (5000,0.0041767005)};
\addlegendentry{Learned shape}
\end{groupplot}
\end{tikzpicture}
\caption{Learning curves over ten paired seeds. Solid curves show the arithmetic mean; shaded bands show one sample standard deviation. Each update uses 1,024 antithetic pairs, or 2,048 observed transitions.}
\label{fig:learning-curves}
\end{figure}

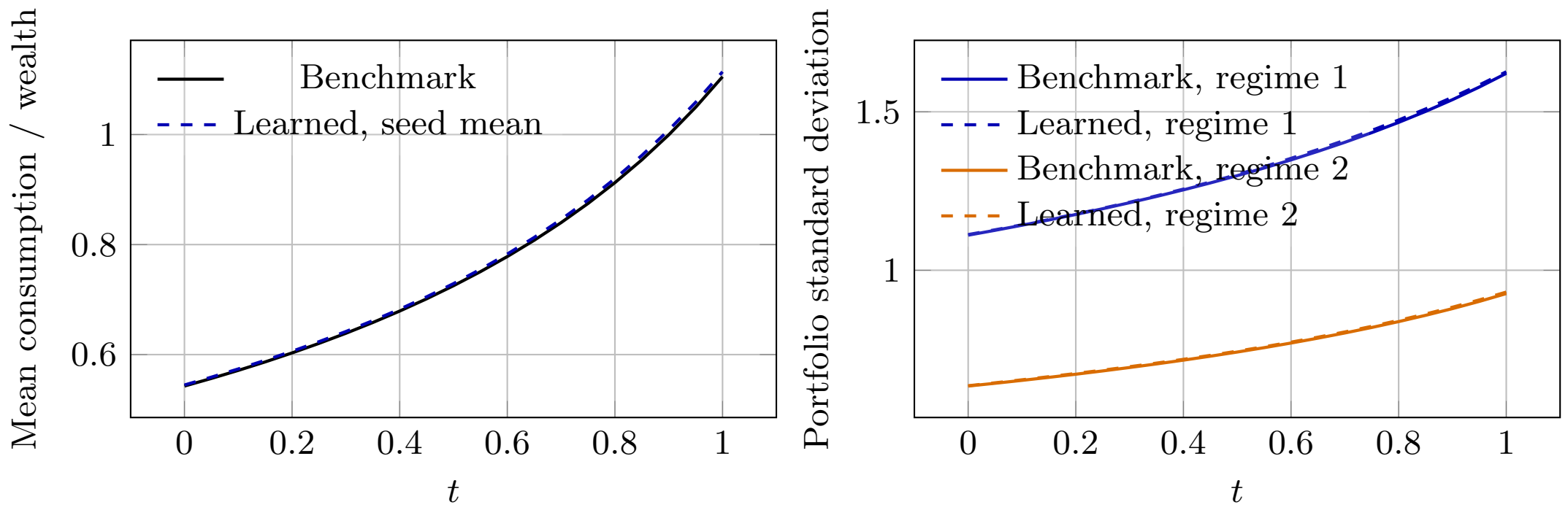
\begin{figure}[tbp]
\centering
\begin{tikzpicture}
\begin{groupplot}[group style={group size=2 by 1,horizontal sep=1.25cm},width=0.45\textwidth,height=5cm,xlabel={$t$},grid=major,legend style={font=\scriptsize,at={(0.02,0.98)},anchor=north west,fill=white,fill opacity=0.15,text opacity=1,draw=none},tick label style={font=\scriptsize},label style={font=\small}]
\nextgroupplot[ylabel={Mean consumption / wealth}]
\addplot[black,thick] coordinates {(0,0.54277005) (0.05,0.55662689) (0.1,0.57124414) (0.15,0.5866861) (0.2,0.60302454) (0.25,0.62033978) (0.3,0.63872205) (0.35,0.65827305) (0.4,0.67910782) (0.45,0.701357) (0.5,0.72516961) (0.55,0.75071634) (0.6,0.77819366) (0.65,0.8078289) (0.7,0.83988653) (0.75,0.8746761) (0.8,0.91256222) (0.85,0.9539774) (0.9,0.99943868) (0.95,1.0495692) (1,1.1051271)};
\addlegendentry{Benchmark}
\addplot[blue!70!black,thick,dashed] coordinates {(0,0.54472705) (0.05,0.5589881) (0.1,0.57390434) (0.15,0.58955964) (0.2,0.60604698) (0.25,0.62346916) (0.3,0.64193975) (0.35,0.66158405) (0.4,0.68254043) (0.45,0.70496181) (0.5,0.72901747) (0.55,0.75489517) (0.6,0.78280369) (0.65,0.81297582) (0.7,0.84567191) (0.75,0.88118402) (0.8,0.91984095) (0.85,0.96201406) (0.9,1.0081243) (0.95,1.0586506) (1,1.1141398)};
\addlegendentry{Learned, seed mean}
\nextgroupplot[
  ylabel={Portfolio standard deviation},
  legend style={font=\scriptsize,at={(0.02,0.98)},anchor=north west,
    fill=white,fill opacity=0.15,text opacity=1,draw=none,
    legend columns=1,cells={anchor=west}}
]
\addplot[blue!70!black,thick] coordinates {(0,1.1106612) (0.05,1.1260281) (0.1,1.1420138) (0.15,1.1586616) (0.2,1.176019) (0.25,1.1941384) (0.3,1.2130779) (0.35,1.232902) (0.4,1.2536825) (0.45,1.2754995) (0.5,1.2984433) (0.55,1.322615) (0.6,1.3481293) (0.65,1.3751165) (0.7,1.4037252) (0.75,1.4341259) (0.8,1.4665155) (0.85,1.5011223) (0.9,1.5382136) (0.95,1.5781036) (1,1.6211656)};
\addlegendentry{Benchmark, regime 1}
\addplot[blue!70!black,thick,dashed] coordinates {(0,1.112677) (0.05,1.1277083) (0.1,1.1434589) (0.15,1.1599775) (0.2,1.1773166) (0.25,1.195532) (0.3,1.2146832) (0.35,1.2348339) (0.4,1.2560519) (0.45,1.2784095) (0.5,1.301984) (0.55,1.3268582) (0.6,1.3531204) (0.65,1.3808654) (0.7,1.4101949) (0.75,1.4412181) (0.8,1.4740527) (0.85,1.5088256) (0.9,1.5456738) (0.95,1.5847456) (1,1.6262019)};
\addlegendentry{Learned, regime 1}
\addplot[orange!85!black,thick] coordinates {(0,0.63466354) (0.05,0.64344461) (0.1,0.65257931) (0.15,0.66209233) (0.2,0.67201083) (0.25,0.68236478) (0.3,0.69318737) (0.35,0.70451542) (0.4,0.71638998) (0.45,0.72885688) (0.5,0.74196758) (0.55,0.75577999) (0.6,0.77035962) (0.65,0.78578088) (0.7,0.8021287) (0.75,0.81950054) (0.8,0.83800885) (0.85,0.8577842) (0.9,0.87897918) (0.95,0.90177348) (1,0.92638031)};
\addlegendentry{Benchmark, regime 2}
\addplot[orange!85!black,thick,dashed] coordinates {(0,0.63563692) (0.05,0.64491876) (0.1,0.65443302) (0.15,0.66422243) (0.2,0.67433242) (0.25,0.68481128) (0.3,0.69571033) (0.35,0.70708415) (0.4,0.71899079) (0.45,0.73149211) (0.5,0.74465415) (0.55,0.7585475) (0.6,0.77324781) (0.65,0.78883634) (0.7,0.80540058) (0.75,0.82303503) (0.8,0.84184195) (0.85,0.86193238) (0.9,0.8834272) (0.95,0.90645837) (1,0.93117033)};
\addlegendentry{Learned, regime 2}
\end{groupplot}
\end{tikzpicture}
\caption{Policy moments for the logarithmic benchmark and the fixed-shape learner after 5,000 updates, averaged over ten seeds. Consumption divided by wealth is independent of the current regime in this particular benchmark.}
\label{fig:policy-moments}
\end{figure}

Figure~\ref{fig:learning-curves} also shows that fixing the shape is not uniformly better at every early checkpoint. The improvement is a reduction in the time to the specified accuracy level, rather than a claim of uniformly dominating learning curves. Figure~\ref{fig:policy-moments} compares the learned policy moments with \eqref{eq:log-benchmark-policy}. In this benchmark the consumption-to-wealth ratio rises as the horizon shortens, and portfolio exploration has smaller standard deviation in the higher-volatility regime. Both effects are recovered by the learner. The reported moment errors use the same time--regime grid as $E_{\mathrm{KL}}$; relative errors divide by the corresponding benchmark moment before taking the root mean square.

Runs used Python 3.12.14, NumPy 2.3.5, and SciPy 1.17.0 on an Intel Xeon Platinum 8573C processor, with one BLAS thread. Timing includes the periodic metric evaluations and excludes the initial benchmark ODE solve. These short wall-clock measurements depend on hardware and implementation; the paired update and transition counts are the primary acceleration measures. 

\subsubsection{Time discretization and scope of the experiment}
To separate mesh error from sampling and optimization error, we compute the population Bellman solution for the transition rule \eqref{eq:numerical-transition} and the left-endpoint reward. Let $K=T/h$ be the integer number of time steps, set $t_k=kh$ for
$k=0,\ldots,K$, and write $P_h=e^{Qh}$ and
$B_k^h=(B_{k,1}^h,B_{k,2}^h)^\top$.
Starting from $A_K^h=e^{-\rho T}$ and $B_K^h=0$, backward optimization gives
\begin{equation}\label{eq:population-recursion}
\begin{aligned}
A_k^h&=A_{k+1}^h+(e^{-\rho t_k}+\lambda)h,\\
B_k^h&=P_hB_{k+1}^h+h\,\mathfrak F(t_k,A_{k+1}^h),
\end{aligned}
\end{equation}
where $\mathfrak F_i(t,a)$ is the right-hand side of \eqref{eq:log-benchmark-F} with each occurrence of $A(t)$ replaced by $a$. The optimal discrete-time consumption distribution has Gamma rate parameter $A_{k+1}^h/(\lambda x)$, and the portfolio distribution has variance $\lambda/(A_{k+1}^h\sigma^2(i))$. These recursions do not involve fitted parameters or Monte Carlo noise. We compare $V^h(t_k,e^y,i)=A_k^h y+B_{k,i}^h$ with the continuous benchmark on $y\in[-1,1]$. Because the difference is affine in $y$, its maximum over that interval is attained at an endpoint.

\begin{table}[tbp]
\centering
\caption{Population mesh error $E_h:=\max_{k,i,|y|\le1}|V^h(t_k,e^y,i)-V(t_k,e^y,i)|$ for the numerical sampling scheme. The observed order is $\log_2(E_h/E_{h/2})$.}
\label{tab:mesh-refinement}
\begin{tabular}{ccc}
\toprule
Step $h$ & Maximum absolute value error & Observed order\\
\midrule
0.0400 & $1.75511\times10^{-2}$ & ---\\
0.0200 & $8.74760\times10^{-3}$ & 1.005\\
0.0100 & $4.36682\times10^{-3}$ & 1.002\\
0.0050 & $2.18167\times10^{-3}$ & 1.001\\
0.0025 & $1.09040\times10^{-3}$ & 1.001\\
\bottomrule
\end{tabular}
\end{table}

Table~\ref{tab:mesh-refinement} exhibits first-order convergence, consistent with the endpoint approximation used here. The learning errors in Table~\ref{tab:learning-results} combine this finite-grid effect with approximation and stochastic-optimization errors. The experiment supports the use of an explicit distributional family and fixed known parameters in this two-regime logarithmic problem. 

\section{Conclusion}\label{sec:conclusion}
This paper develops a distribution-informed approach to
entropy-regularized reinforcement learning for regime-switching
stochastic control. The action domain and the action dependence
of the Hamiltonian determine explicit exploratory policy families
and the parameter restrictions required for normalization.
These relations provide a distributional structure prior for
learning: they specify the policy support, connect the
action-dependent coefficients with the normalizing term, and
identify parameters that can be fixed in advance. Incorporating
this information into a shared policy and $q$-function
parameterization avoids a separate action-space normalization
step and reduces the number of quantities estimated from data.
The remaining parameter functions can be learned without
supplying the numerical drift, volatility, or regime-transition
coefficients, provided that the relevant structural action
dependence is known.

The consumption-investment application illustrates both the
analytical and computational roles of this prior. For any smooth
finite-valued solution of the exploratory HJB equation, finiteness
of the optimized Hamiltonian implies positive marginal value
and strictly negative wealth curvature, ensuring that the
Gamma consumption and Gaussian investment densities are
well defined. A logarithmic terminal-utility benchmark provides
an explicit wealth coefficient and a linear system for the
regime-dependent value offsets. It also identifies a consumption
shape parameter that need not be estimated. In ten paired
numerical runs, fixing this parameter reduces the mean update
count required to reach the prescribed policy-error threshold
by 12.6\%, while maintaining comparable final accuracy.
The comparison demonstrates a concrete benefit of incorporating
known distributional information within the same policy family.
A separate mesh study identifies first-order error from the
sampled implementation.

The usefulness of the distributional prior depends on correctly
identifying the policy structure and verifying the admissibility
conditions for the chosen parameter functions. The present
numerical evidence concerns an observed-regime logarithmic
benchmark, and convergence guarantees for the general
parameterized learning scheme remain an open question.
Further work may consider nonlogarithmic preferences, partially
observed regimes, and coupled actions, as well as the effect
of misspecified distributional information. Error bounds
combining approximation, sampling, optimization, and time
discretization would further clarify when the reduction in
learned parameters leads to improved learning efficiency.

\section*{Declarations}

\subsection*{Author contributions}
All authors contributed equally to this work. All authors read and
approved the final manuscript.

\subsection*{Funding}
This work was supported by the National Key R\&D Program of China
(2022YFA1007900), the National Natural Science Foundation of China
(12671572), Fundamental Research Funds for the Central Universities
(2024QKT008, 2026QKT006, 2025ECNU-WLJC006), and Shanghai Sci-tech
Co-research Program (25HB2702200). The funders had no role in the
study design, analysis, manuscript preparation, or decision to
submit the article for publication.

\subsection*{Declaration of competing interest}
The authors declare that they have no known competing financial
interests or personal relationships that could have appeared to
influence the work reported in this paper.

\subsection*{Declaration of generative AI use}
During the preparation of this manuscript, the authors used ChatGPT
(OpenAI) to improve the language and readability and to assist with
\LaTeX{} formatting. After using this tool, the authors reviewed and
edited the content as needed and take full responsibility for the
content of the published article.

\begin{appendices}
\section{Derivation of the regime-switching exploratory SDE}
\label{Appendix  exploratory SDE}
\setcounter{equation}{0}
\renewcommand{\theequation}{A.\arabic{equation}}

Fix $\boldsymbol{\pi}\in\boldsymbol{\Pi}$ and a test function
$f$ such that
$f(\cdot,\cdot,i)\in C^{1,2}([0,T)\times\mathbb{R}^d)$
with compact support for each $i\in\mathcal{S}$.
For a fixed action $a\in\mathcal{A}$, the regime-switching
It\^{o} formula gives the extended generator
\begin{equation}\label{eq:appA:controlled_generator}
\begin{aligned}
\mathcal{L}^{a}f(s,x,i)
={}&f_s(s,x,i)
+b(s,x,i;a)\circ f_x(s,x,i)\\
&+\frac12\sigma^2(s,x,i;a)\circ f_{xx}(s,x,i)
+\sum_{j=1}^{m}q_{ij}f(s,x,j).
\end{aligned}
\end{equation}

At a fixed state $(s,x,i)$, the exploratory formulation averages
the controlled generator with respect to
$\boldsymbol\pi(\cdot\mid s,x,i)$. This defines the relaxed generator
\begin{equation}\label{eq:appA:averaging}
\begin{aligned}
\mathcal{L}^{\boldsymbol{\pi}}f(s,x,i)
&=\int_{\mathcal{A}}\mathcal{L}^{a}f(s,x,i)
\boldsymbol{\pi}(a\mid s,x,i)\,da\\
&=f_s(s,x,i)
+\tilde b(s,x,i;\boldsymbol{\pi})\circ f_x(s,x,i)\\
&\quad+\frac12\tilde\sigma^2(s,x,i;\boldsymbol{\pi})
\circ f_{xx}(s,x,i)
+\sum_{j=1}^{m}q_{ij}f(s,x,j),
\end{aligned}
\end{equation}
where
\begin{equation}\label{eq:appA:averaged_coefficients}
\begin{aligned}
\tilde b(s,x,i;\boldsymbol{\pi})
&=\int_{\mathcal{A}}b(s,x,i;a)
\boldsymbol{\pi}(a\mid s,x,i)\,da,\\
\tilde\sigma(s,x,i;\boldsymbol{\pi})
&=\left\{
\int_{\mathcal{A}}\sigma^2(s,x,i;a)
\boldsymbol{\pi}(a\mid s,x,i)\,da
\right\}^{1/2}.
\end{aligned}
\end{equation}
The coefficient integrability in
Definition~\ref{def:admissible_policy} makes the action integral of
the controlled generator finite at each state. Linearity of integration
then yields \eqref{eq:appA:averaging}. The switching term remains
unchanged because $Q$ is independent of the action and the policy
density integrates to one. 

The matrix inside the square root is symmetric and positive
semidefinite. Thus, $\tilde\sigma$ is well-defined and satisfies
\[
\tilde\sigma\tilde\sigma^\top
=\int_{\mathcal{A}}\sigma\sigma^\top(s,x,i;a)
\boldsymbol{\pi}(a\mid s,x,i)\,da.
\]
Consequently, an SDE representation of
$\mathcal{L}^{\boldsymbol{\pi}}$ is
\begin{equation}\label{eq:appA:exploratory_sde}
\left\{
\begin{aligned}
dX_s
={}&\tilde b(s,X_s,\theta_s;\boldsymbol{\pi}_s)\,ds
+\tilde\sigma(s,X_s,\theta_s;\boldsymbol{\pi}_s)\,d\mathcal{W}_s,\\
X_t={}&x,\qquad \theta_t=i,
\end{aligned}
\right.
\end{equation}
where
$\boldsymbol{\pi}_s=\boldsymbol{\pi}(\cdot\mid s,X_s,\theta_s)$
and $\mathcal{W}$ is a $d$-dimensional Brownian motion independent of
$\theta$ on a suitable probability space.
Applying It\^{o}'s formula verifies that this SDE has the
averaged generator in \eqref{eq:appA:averaging}, giving
the exploratory dynamics \eqref{SDE}.

\section{Proof of Proposition \ref{Proposition 1}}\label{Appendix B}
\setcounter{equation}{0}
\renewcommand{\theequation}{B.\arabic{equation}}
\begin{proof}
Fix $\boldsymbol\pi\in\boldsymbol\Pi$ and use the averaged coefficients
$\tilde b$ and $\tilde\sigma$ from
Definition~\ref{def:admissible_policy}. Their action integrals are finite
by condition~$(\romannumeral 1)$, and the principal square root of the
averaged covariance is well-defined. Condition~$(\romannumeral 5)$ gives
\[
\begin{aligned}
&|\tilde b(s,x,i;\boldsymbol\pi)-\tilde b(s,y,i;\boldsymbol\pi)|
+|\tilde\sigma(s,x,i;\boldsymbol\pi)-\tilde\sigma(s,y,i;\boldsymbol\pi)|
\leq\ell_{K,\boldsymbol\pi}|x-y|,\\
&|\tilde b(s,x,i;\boldsymbol\pi)|+|\tilde\sigma(s,x,i;\boldsymbol\pi)|
\leq c_{\boldsymbol\pi}(1+|x|),
\end{aligned}
\]
where the first inequality holds for $x,y\in\mathbb R_K^d$ and both
bounds are uniform in $s$ and $i$.

For each fixed regime, these are the local Lipschitz and linear growth
conditions for a unique nonexplosive strong diffusion. The finite-state
chain has bounded jump intensities and hence only finitely many jumps
on $[t,T]$ almost surely. Construct the solution successively between
its jump times, using the same independent Brownian motion
$\mathcal W$ and the state reached at the preceding jump. This gives a
strong solution on the whole interval. Pathwise uniqueness on each
interval gives pathwise uniqueness of the interlaced solution
(see \citet{mao2006stochastic}).
\end{proof}

\section{Proof of Proposition \ref{Proposition 2}}\label{Appendix C}
\setcounter{equation}{0}
\renewcommand{\theequation}{C.\arabic{equation}}
\begin{proof}
Fix an admissible policy $\boldsymbol\pi$ and $p\geq2$.
By \eqref{eq:averaged_linear_growth}, there exists
$c_{\boldsymbol\pi}>0$, independent of $t,x,i$, such that
\[
|\tilde b(s,y,j;\boldsymbol\pi)|+|\tilde\sigma(s,y,j;\boldsymbol\pi)|
\leq c_{\boldsymbol\pi}(1+|y|).
\]
Let $\rho_N:=T\wedge\inf\{s\geq t:|X_s^{\boldsymbol\pi}|\geq N\}$.
Apply the integral form of \eqref{SDE} to the stopped process.
The elementary inequality
$|u+v+w|^p\leq3^{p-1}(|u|^p+|v|^p+|w|^p)$, H\"older's inequality,
and the Burkholder--Davis--Gundy inequality give
\[
\begin{aligned}
&\E_{t,x,i}\sup_{t\leq u\leq s}
 \left|\int_t^{u\wedge\rho_N}\tilde b(r,X_r^{\boldsymbol\pi},\theta_r;\boldsymbol\pi)\,dr\right|^p\\
&\qquad\leq T^{p-1}\E_{t,x,i}
 \int_t^{s\wedge\rho_N}|\tilde b(r,X_r^{\boldsymbol\pi},\theta_r;\boldsymbol\pi)|^p\,dr,\\
&\E_{t,x,i}\sup_{t\leq u\leq s}
 \left|\int_t^{u\wedge\rho_N}\tilde\sigma(r,X_r^{\boldsymbol\pi},\theta_r;\boldsymbol\pi)\,d\mathcal W_r\right|^p\\
&\qquad\leq C_p^{\mathrm{BDG}}T^{p/2-1}\E_{t,x,i}
 \int_t^{s\wedge\rho_N}|\tilde\sigma(r,X_r^{\boldsymbol\pi},\theta_r;\boldsymbol\pi)|^p\,dr.
\end{aligned}
\]
Define $F_N(s):=\E_{t,x,i}
\sup_{t\leq u\leq s}|X_{u\wedge\rho_N}^{\boldsymbol\pi}|^p$.
Combining the preceding estimates with linear growth yields a constant
$C_2=C_2(p,T,c_{\boldsymbol\pi})>0$, independent of $t,x,i,N$, such that
\[
F_N(s)\leq C_2\left(1+|x|^p+\int_t^s F_N(u)\,du\right).
\]
Gronwall's inequality therefore gives
\[
F_N(T)\leq C_2(1+|x|^p)e^{C_2(T-t)}
\leq C_2e^{C_2T}(1+|x|^p).
\]
The solution is nonexplosive by Proposition~\ref{Proposition 1}.
Letting $N\to\infty$ and applying Fatou's lemma proves
\[
\E_{t,x,i}\left[\sup_{t\leq s\leq T}|X_s^{\boldsymbol\pi}|^p\right]
\leq C_{p,\boldsymbol\pi}(1+|x|^p),
\quad C_{p,\boldsymbol\pi}:=C_2e^{C_2T}.
\]
This constant is independent of the initial time, state, and regime.
Its possible dependence on the policy is through the averaged linear
growth bound. It can be chosen uniformly over any policy subclass for
which $c_{\boldsymbol\pi}$ has a common upper bound.
\end{proof}

\section{Proof of Proposition
\ref{proposition increasing and concave Merton}}\label{Appendix D}
\setcounter{equation}{0}
\renewcommand{\theequation}{D.\arabic{equation}}

\begin{proof}
Fix $(t,x,i)\in[0,T)\times(0,\infty)\times\mathcal S$.
All wealth derivatives below are evaluated at $(t,x,i)$.
Since $V_t(t,x,i)$ and the regime-switching term
$\sum_{j=1}^m q_{ij}V(t,x,j)$ are finite, the pointwise
supremum in \eqref{HJB equation} must also be finite.
By the entropy variational formula, applied first on bounded
action sets and then by exhaustion of $\mathcal A$, this requires
\begin{equation}\label{eq:finite-local-hamiltonian}
0<
\int_{\mathcal A}
\exp\left(
\frac{1}{\lambda}
H(t,x,i,V_x,V_{xx};c,u)
\right)\,dc\,du
<\infty.
\end{equation}

For the consumption-investment model, the Hamiltonian is
\[
\begin{aligned}
H(t,x,i,V_x,V_{xx};c,u)
={}&e^{-\rho t}\ln(c)-cV_x\\
&+\frac12 x^2\sigma^2(i)V_{xx}u^2
+x(\mu(i)-R(i))V_xu+xR(i)V_x.
\end{aligned}
\]
The last term is finite and independent of the actions.
The remaining terms separate into a consumption component
and an investment component. Since their exponential factors
are strictly positive, \eqref{eq:finite-local-hamiltonian}
implies that both
\[
\int_0^\infty
c^{e^{-\rho t}/\lambda}
\exp\left(-\frac{V_x}{\lambda}c\right)\,dc
<\infty
\]
and
\[
\int_{\mathbb R}
\exp\left(
\frac{1}{\lambda}
\left[
\frac12 x^2\sigma^2(i)V_{xx}u^2
+x(\mu(i)-R(i))V_xu
\right]
\right)\,du
<\infty.
\]

First, suppose that $V_x\leq0$. Since
$e^{-\rho t}/\lambda>0$, the consumption integral satisfies
\[
\int_0^\infty
c^{e^{-\rho t}/\lambda}
\exp\left(-\frac{V_x}{\lambda}c\right)\,dc
\geq
\int_1^\infty c^{e^{-\rho t}/\lambda}\,dc
=\infty,
\]
which is a contradiction. Therefore, $V_x>0$.

Next, suppose that $V_{xx}>0$. Since $x>0$ and
$\sigma(i)>0$, the quadratic coefficient in the investment
exponent is strictly positive. The quadratic term dominates
the linear term as $|u|\to\infty$, so the investment
integral diverges. If $V_{xx}=0$, the integrand is the
exponential of a linear function of $u$, or a positive
constant when the linear coefficient vanishes. Its integral
over $\mathbb R$ is infinite in either case. Hence,
$V_{xx}<0$.

Since $(t,x,i)$ was arbitrary, these strict inequalities hold
throughout $[0,T)\times(0,\infty)\times\mathcal S$.
Thus, for every fixed $t<T$ and $i\in\mathcal S$,
$V(t,\cdot,i)$ is strictly increasing and strictly concave.
\end{proof}
\end{appendices}

%%===========================================================================================%%
%% If you are submitting to one of the Nature Portfolio journals, using the eJP submission   %%
%% system, please include the references within the manuscript file itself. You may do this  %%
%% by copying the reference list from your .bbl file, paste it into the main manuscript .tex %%
%% file, and delete the associated \verb+\bibliography+ commands.                            %%
%%===========================================================================================%%

%% if required, the content of .bbl file can be included here once bbl is generated
%%\input sn-article.bbl

\end{document}